\documentclass{article}

\usepackage[T1]{fontenc}
\usepackage{microtype}
\usepackage{graphicx}
\usepackage{subcaption}
\usepackage{booktabs}
\usepackage{tabularx}
\usepackage{multirow}
\usepackage{xcolor}
\usepackage[normalem]{ulem}  % \sout for \del
\usepackage{hyperref}   

\usepackage[preprint]{icml2026}

\usepackage{amsmath}
\usepackage{amssymb}
\usepackage{mathtools}
\usepackage[capitalize,noabbrev]{cleveref}

\graphicspath{{figures/}}

\providecommand{\R}{\mathbb{R}}
\newcommand{\taskicon}[1]{\csname taskicon#1\endcsname}

\icmltitlerunning{Learning Patient Representations from Partial Observations}

\begin{document}

\twocolumn[
  \icmltitle{Learning Patient Representations from Partial Observations\texorpdfstring{\\}{}
     through Reconstruction and Latent Prediction}

  % Author information is removed by the style file in the blind version.
  \icmlsetsymbol{senior}{$\dagger$}
  \begin{icmlauthorlist}
    \icmlauthor{Max Kr\"ahenmann}{basel}
    \icmlauthor{Andrea Agostini}{eth}
    \icmlauthor{Simon B\"ohi}{basel}
    \icmlauthor{Moritz Vandenhirtz}{eth}
    % last authors on a line of their own (no break in the blind version)
    \ificmlshowauthors\\\fi
    \icmlauthor{Julia E.\ Vogt}{senior,eth}
    \icmlauthor{Thomas M.\ Sutter}{senior,eth}
    \icmlauthor{Ece Ozkan}{senior,basel}
  \end{icmlauthorlist}

  \icmlaffiliation{basel}{Department of Biomedical Engineering, University of Basel, Switzerland}
  \icmlaffiliation{eth}{Department of Computer Science, ETH Zurich, Switzerland}

  \icmlcorrespondingauthor{Max Kr\"ahenmann}{max.kraehenmann@unibas.ch}

  \icmlkeywords{self-supervised learning, electronic health records, irregular time series, masked autoencoders, joint-embedding predictive architectures, intensive care}

  \vskip 0.3in
]

% Give the unnumbered affiliations footnote an anchor name (hyperref otherwise warns about an empty anchor)
\makeatletter\def\Hy@footnote@currentHref{Hfootnote.affiliations}\makeatother
\printAffiliationsAndNotice{\textsuperscript{$\dagger$}Shared senior authorship. }

\begin{abstract}
Intensive-care records only provide partial observations of a patient: measurements are sparse, irregular and noisy, and their choice and timing follow clinical practices that vary across clinicians and hospitals. Moreover, clinical risk and diagnosis predictions are most useful early on in a stay, and labels for these tasks are scarce and costly to acquire, as many require expert review. Self-supervised pretraining can learn the structure of large unlabelled cohorts and yield reusable representations, but its dominant objectives, masked reconstruction and forecasting, operate in input space: they supervise the representation only through errors on individual measured values, which incentivizes retaining low-level detail instead of semantics. We introduce \textbf{GLIMPSE} (\textbf{G}lobal \textbf{L}atent \textbf{I}nference and \textbf{M}asked \textbf{P}rediction for \textbf{S}parse \textbf{E}HR), a sparse masked autoencoder that encodes only observed measurements and adds a latent prediction objective whose target lies largely outside its context: it predicts the representation of a whole stay from a glimpse of it. On two adult and one paediatric cohort, GLIMPSE learns better representations than state-of-the-art self-supervised encoders for mortality, length-of-stay and diagnosis prediction, and its advantage over these encoders grows when labels are scarce and holds when observations are limited.
\end{abstract}

% Previous abstract:
% \begin{abstract}
% Self-supervised pretraining promises patient representations that can be reused across clinical tasks, but intensive-care records are sparse, irregularly sampled and shaped by clinical practice. Masked reconstruction, the dominant pretraining objective for such records, supervises the representation only through errors on individual measured values. We introduce \textbf{GLIMPSE} (\textbf{G}lobal \textbf{L}atent \textbf{I}nference and \textbf{M}asked \textbf{P}rediction for \textbf{S}parse \textbf{E}HR), a sparse masked autoencoder that encodes only observed measurements and adds a latent prediction objective: it predicts the representation of the whole stay from the representation of a glimpse of it. On two adult cohorts and one paediatric cohort, GLIMPSE yields representations that in linear probing perform best among state-of-the-art baselines on downstream tasks such as mortality-, ICD code- and length of stay prediction, and keep their advantage on average on full finetuning. Furthermore, GLIMPSE is label-efficient, outperforming baselines when labels are scarce; observation-efficient, remaining the best encoder when observations are limited; and parameter-efficient, with a variant of only 154k parameters outperforming models hundreds of times bigger.
% \end{abstract}

\section{Introduction}
\label{sec:introduction}

Patients in intensive care units (ICUs) are monitored closely, and the measurements taken during their stay are stored in electronic health records (EHRs). These records make it possible to learn models whose predictions support clinical decisions. Such predictions can improve survival through the early recognition of conditions such as sepsis \citep{seymour2017time,reyna2020early}, reduce missed diagnoses \citep{winters2012diagnostic}, and support triage and bed planning \citep{harutyunyan2019multitask}. Building such models is, however, challenging.

Firstly, the EHR observes the patient only partially. Vital signs and laboratory results are sparse, with missingness rates of over 80\% in common ICU benchmarks \citep{che2018grud,zhang2022raindrop}, irregularly timed \citep{horn2020seft,shukla2021mtan} and noisy \citep{maslove2016errors}, and which variables are measured, and when, reflects the clinician's assessment \citep{che2018grud,agniel2018biases}. Whether a patient is recovering or deteriorating must be inferred from these scattered measurements, and a prediction is most useful if done early on \citep{silva2012predicting}. Secondly, every endpoint needs labels of its own, diagnoses require expert review, and many hospitals contribute only a few hundred stays \citep{pollard2018eicu,mehrjou2022fedicu}; the median ICU machine-learning study analyses around 488 patients \citep{shillan2019use}. Self-supervised pretraining can ease both difficulties: it learns a representation of the record once, from unlabelled stays, and this representation can then be adapted to each endpoint with few labels.

How well it does so depends on the pretraining objective and on the model's cost. Masked autoencoders such as DuETT and AID-MAE reconstruct held-out measurements, and STraTS forecasts  the next two hours \citep{labach2023duett,xiang2026aidmae,tipirneni2022strats}. All three are scored value by value on the raw measurements, which pushes the representation to encode fine-grained detail of each measurement. Joint-embedding predictive architectures (JEPA) have shown promise for images, video and time series \citep{assran2023ijepa,bardes2024vjepa,ennadir2024tsjepa}. They predict learned representations rather than the input, thereby reducing the amount of noise they encode \citep{lecun2022path}. This preference for abstract features, however, comes at the cost of fine-grained detail on which many diagnoses rest \citep{littwin2024jepa}. Tabular foundation models, in turn, have achieved strong performance on tabular data \citep{hollmann2025tabpfn,qu2026tabiclv2,kong2026tabfmzeroshotfoundationmodel}, but they are large models, whereas hospitals often have to train and run models on local infrastructure (Section~\ref{sec:results-cost}).

We therefore propose \textbf{GLIMPSE}, a new representation learning technique that leverages a small sparse encoder that embeds only the observed measurements and, alongside reconstructing masked values, predicts the representation of the whole stay from a short window of it. Unlike prior latent objectives for EHRs, which predict masked values within the window or the next hour \citep{yu2024smart,yang2026clinjepa}, GLIMPSE predicts a target beyond its input window, aligning representations of views with little overlap.

We evaluate GLIMPSE on mortality, length-of-stay and diagnosis prediction in two adult cohorts and a paediatric one, a population that is rarely studied: none of 53 recently reviewed EHR foundation models was pretrained exclusively on children \citep{guo2026systematic}. It gives the best representations on average, with a lead that grows when labels are scarce and holds when observations are limited, and outperforms encoders up to 9 times its size. In summary, our contributions are the following:

\begin{itemize}
    \item A small sparse masked autoencoder for EHRs that leverages the structure of the time--variable grid while encoding only its observed entries.
    \item A latent objective that predicts the whole stay's representation from a short window and, combined with value reconstruction, improves performance.
    \item A comparison of recent self-supervised EHR encoders and tabular foundation models on rarely evaluated paediatric intensive care data alongside two adult cohorts.
\end{itemize}

\section{Related Work}
\label{sec:related-work}

\paragraph{Encoding irregular ICU records.}
Coded EHRs represent patient histories as sequences of diagnosis, procedure and visit tokens \citep{li2020behrt,rasmy2021medbert,steinberg2024motor}, but lack the fine-grained physiological measurements needed to follow rapidly evolving patient states in intensive care. Time--variable grids align these measurements across variables and time: GRU-D \citep{che2018grud} fills the missing cells by imputation with a learned decay, and DuETT \citep{labach2023duett} shows that attention alternating over the time and variable axes of a binned grid is a promising way to exploit this structure, but computes over its mostly empty cells. Other models encode only the observed measurements: SeFT \citep{horn2020seft} aggregates them as a set, without modelling pairwise interactions; mTAN \citep{shukla2021mtan} attends over them in continuous time; STraTS \citep{tipirneni2022strats} applies self-attention to (time, variable, value) triplets, at a memory cost quadratic in their number; and AID-MAE \citep{xiang2026aidmae} groups them into daily records, which limits interactions within a day and between distant days.

\paragraph{Reconstruction and forecasting.}
Most self-supervised objectives for EHRs predict values or events held out from the input. Masked autoencoders reconstruct masked measurements from the visible ones \citep{he2022masked}, as in DuETT and AID-MAE, and TransEHR \citep{xu2023transehr} adds the detection of replaced triplets. Forecasting objectives predict what follows the input instead: the values of the next two hours in STraTS, the time until measurements cross varying thresholds in ICareFM \citep{burger2025icarefm}, and the time to future events in MOTOR \citep{steinberg2024motor}, while BERT-style models of coded EHRs predict masked codes \citep{li2020behrt,rasmy2021medbert}. All these objectives supervise the representation only through errors on individual values or events, which can often be estimated from neighbouring observations \citep{cheng2023timemae} and include measurement noise \citep{littwin2024jepa}.

\paragraph{Predicting representations.}
A second family predicts learned representations rather than the input, so that the model need not reproduce unpredictable low-level detail. BYOL \citep{grill2020byol} and data2vec \citep{baevski2022data2vec} regress the representations of an exponential-moving-average (EMA) teacher, and multi-crop self-distillation \citep{caron2021dino,oquab2024dinov2} predicts the representation of a global view from local crops. iBOT \citep{zhou2022ibot}, data2vec~2.0 \citep{baevski2023data2vec2} and JEPAs \citep{lecun2022path,assran2023ijepa} predict the representations of masked inputs \citep{bardes2024vjepa,ennadir2024tsjepa}. In EHRs, SMART \citep{yu2024smart} regresses EMA-teacher latents of masked values on an hourly grid, and Clin-JEPA \citep{yang2026clinjepa} predicts the next hour's latent state of text-serialised ICU stays. These methods use latent targets in place of reconstruction rather than alongside it, and can therefore lose fine-grained detail \citep{littwin2024jepa}; in EHRs, their targets are also limited to the same or the next time step.

\paragraph{Tabular foundation models.}
TabPFN \citep{hollmann2025tabpfn}, TabICL \citep{qu2026tabiclv2} and TabFM \citep{kong2026tabfmzeroshotfoundationmodel} are pretrained on large, often synthetic, collections of tables and predict on a new dataset in context from its labelled rows, without gradient updates. They are strong on small tables, but applying them to clinical time series requires encoding each stay, with hundreds to thousands of irregular measurements, into a fixed set of features.

\section{Method}
\label{sec:method}

GLIMPSE is a self-supervised Transformer-based encoder that learns the representation of a patient stay from irregularly sampled measurements associated with it. Its pipeline is composed of three main steps. First, every observed measurement becomes a token that carries its value, variable and time, and missing entries are excluded (Section~\ref{sec:data}). Second, a single Transformer encoder is pretrained on these tokens with two complementary objectives (Section~\ref{sec:objectives}): masked reconstruction recovers hidden values from the visible ones, and latent prediction predicts the representation of the whole stay from that of a short window of it. Third, only the encoder is kept after pretraining, and it is able to encode an arbitrary number of hours of a stay into the representation used for prediction (Section~\ref{sec:downstream}).

\begin{figure*}[ht]
    \centering
    \includegraphics[width=\textwidth]{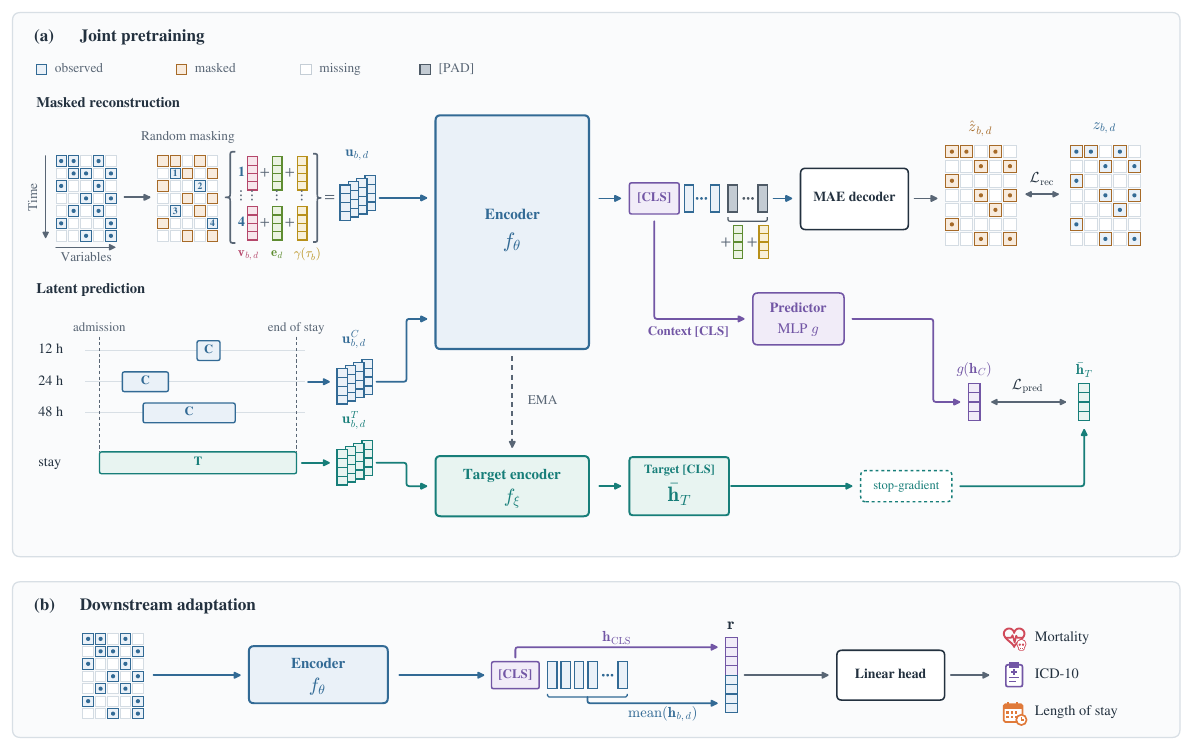}
    \caption{Overview of GLIMPSE. \textbf{(a)} Pretraining. Only the observed measurements kept by random masking are embedded by value, variable and time and encoded; the decoder reconstructs the masked values from \texttt{[PAD]} mask tokens, which fill every masked or missing entry ($\mathcal{L}_{\mathrm{rec}}$). A context window of 12, 24 or 48 hours and the whole stay, unmasked and in 12 bins each, pass through the encoder and its EMA target encoder (gray dashed arrow), respectively, and a predictor maps the context's \texttt{[CLS]} output to the whole stay's ($\mathcal{L}_{\mathrm{pred}}$, no gradient to the target). \textbf{(b)} Downstream, a linear head reads the concatenated \texttt{[CLS]} and mean value-token outputs.}
    \label{fig:overview}
\end{figure*}

\subsection{Data}
\label{sec:data}
\paragraph{Input and missingness.}
A hospital stay $\mathcal{S}=\{(t_i, d_i, v_i)\}_{i=1}^{N}$ consists of $N$ irregularly sampled measurements. Each one is a value $v_i$ of one of $D$ clinical variables $d_i$, recorded $t_i$ hours after admission. We divide the stay into hourly bins counted from admission, which preserves fine-grained events and tolerates inaccurate timestamps. We keep the latest measurement $x_{b,d}$ of variable $d$ in bin $b$; volumes like urine output are instead summed. This yields a sparse matrix $X\in \R^{B\times D}$, where $B$ is the length of stay in hours. Most of its entries are empty, and not at random: whether a test is ordered or not reflects the clinician's assessment, so the pattern of measurements is itself informative~\citep{che2018grud,agniel2018biases}. We treat missingness as intrinsic to the data~\citep{xiang2026aidmae}. The missingness mask $M\in\{0,1\}^{B\times D}$, with $m_{b,d}=1$ if variable $d$ was measured in bin $b$, is fixed by the data. During pretraining, a random mask $A\in\{0,1\}^{B\times D}$ hides a random 75\% of the observed entries, as in MAE~\citep{he2022masked}, keeping at least one visible. Only the entries that are both observed and kept become tokens,
\begin{equation}
    X_{M \odot A} = \{\, x_{b,d} : m_{b,d}=1,\ a_{b,d}=1 \,\},
    \label{eq:visible}
\end{equation}
so missing entries are never tokenized, imputed or reconstructed, and the cost of the encoder follows the number of measurements rather than the size of the grid.

\paragraph{Tokenization.}
The flattened grid has no positions, and a Transformer is invariant to the order of its inputs, so each token carries its value, variable and time as a sum of three embeddings,
\begin{equation}
    \mathbf{u}_{b,d} = \mathbf{w}\, z_{b,d} + \mathbf{c} + \mathbf{e}_d + \gamma(\tau_b).
    \label{eq:token}
\end{equation}
Clipping extreme measurements is a common step in EHR preprocessing~\citep{rocheteau2021tpc,labach2023duett}. We use each variable's 5th and 95th percentiles as bounds~\citep{xiang2026aidmae}, then standardize the resulting values to obtain $z_{b,d}$. The cutoffs, means and standard deviations are estimated from the training stays. We embed $z_{b,d}$ with $\mathbf{w}, \mathbf{c} \in \R^{E}$, where $E$ is the embedding dimension. Each variable has a learned embedding $\mathbf{e}_d \in \R^{E}$, and the bin's time $\tau_b$ in hours since admission is embedded with fixed sinusoids $\gamma$~\citep{vaswani2017attention} whose periods range from two hours to 30 days (see Appendix~\ref{app:method-details} for more details). Because time is an explicit coordinate rather than a sequence position, the encoder can process tokens from any subset of bins, of any length and resolution.

\subsection{Architecture and pretraining objectives}
\label{sec:objectives}

A single Transformer encoder $f_\theta$ is trained by two objectives, each through a light module that serves only pretraining (Figure~\ref{fig:overview}): a decoder reconstructs masked values, and a predictor, together with a momentum target encoder, predicts the representation of the whole stay from that of a part of it.

\paragraph{Encoder.} A learned \texttt{[CLS]} token \citep{devlin2019bert} is prepended to the tokens of Eq.~\eqref{eq:token}. We denote the encoding of it as $\mathbf{h}_{\mathrm{CLS}}$ and the encoding of token $\mathbf{u}_{b,d}$ as $\mathbf{h}_{b,d}$. As in MAE, the encoder never sees mask tokens or missing entries.

\paragraph{Masked reconstruction.} We use standard MAE~\citep{he2022masked} on a random crop of at most 48 bins of the stay. A Transformer decoder receives the encoder outputs and a mask token for every masked or missing entry, and predicts the standardized values $\hat z_{b,d}$, trained with the mean squared error over the observed, masked entries $\Omega_{\mathrm{mask}} = \{(b,d) : m_{b,d}=1,\ a_{b,d}=0\}$,
\begin{equation}
    \mathcal{L}_{\mathrm{rec}} = \frac{1}{|\Omega_{\mathrm{mask}}|} \sum_{(b,d)\in\Omega_{\mathrm{mask}}} \big(\hat z_{b,d} - z_{b,d}\big)^2 .
    \label{eq:rec}
\end{equation}

\paragraph{Latent prediction.} Alongside reconstruction, our latent prediction path predicts the representation of the whole stay from that of a glimpse of it. For each stay, we sample a context length $s \in \{12, 24, 48\}$ hours and a start hour $a$ such that the context $C=[a, a+s)$ contains at least one measurement. The target $T$ is the whole stay. To predict it, the model is encouraged to align representations of short windows, varying in size and placement, to the one of the whole stay, potentially discarding measurement noise. Every window is divided into $K=12$ bins, hourly for a 12-hour context and coarser for longer windows, so that its cost does not grow with its duration (Appendix~\ref{app:method-details}).

The shared encoder $f_\theta$ embeds the unmasked context's \texttt{[CLS]} token, $\mathbf{h}_C = \mathbf{h}_{\mathrm{CLS}}(C)$. The target $\bar{\mathbf{h}}_T$ is the \texttt{[CLS]} output of a target encoder $f_\xi$ on $T$ whose weights do not get trained and are instead an exponential moving average of the encoder's, $\xi \leftarrow m\,\xi + (1-m)\,\theta$ with $m=0.999$~\citep{baevski2022data2vec}. A two-layer MLP predictor $g$ is trained with the cosine distance~\citep{grill2020byol} to predict $T$ from $C$,
\begin{equation}
    \mathcal{L}_{\mathrm{pred}} = 1 - \cos\!\big(g(\mathbf{h}_C),\, \bar{\mathbf{h}}_T\big).
    \label{eq:pred}
\end{equation}

\paragraph{Joint training.} The pretraining loss is $\mathcal{L} = \mathcal{L}_{\mathrm{rec}} + \lambda_{\mathrm{pred}}\, \mathcal{L}_{\mathrm{pred}}$. Reconstruction keeps the encoder tied to the observed values, while latent prediction makes the representation of a few hours agree with that of the whole stay. With $\lambda_{\mathrm{pred}}=0$, the model is a plain sparse masked autoencoder, which we refer to as GLIMPSE-Rec. Conversely, GLIMPSE-Pred is trained with latent prediction alone, $\mathcal{L} = \mathcal{L}_{\mathrm{pred}}$, without the decoder.

\subsection{Downstream adaptation}
\label{sec:downstream}
After pretraining, the decoder, the predictor and the target encoder are discarded, and the encoder receives all observed values of the first $H$ hours of the stay, without masking. As in established ICU benchmarks~\citep{silva2012predicting,harutyunyan2019multitask}, we choose $H=48$ in our evaluations. We do this for two reasons: predictions are most useful early in the stay, and measurements near its end, for example shortly before death, are closely correlated with the outcome (Section~\ref{sec:exp-setting}). Since latent prediction acts on the \texttt{[CLS]} output while reconstruction can be met by the value-token outputs, the patient representation concatenates both (Appendix~\ref{app:readout}),
\begin{gather}
    \mathbf{r} = \Big[\, \mathbf{h}_{\mathrm{CLS}} \,;\, \tfrac{1}{|\Omega|} \textstyle\sum_{(b,d)\in\Omega} \mathbf{h}_{b,d} \,\Big] \in \R^{2E},
    \label{eq:readout} \\
    \Omega = \{(b,d) : m_{b,d}=1\}. \nonumber
\end{gather}
GLIMPSE-Rec also benefits from this readout. A linear head on $\mathbf{r}$ is trained either on the frozen encoder (linear probing) or jointly with the encoder (full finetuning).

\section{Results}
\label{sec:results}

After describing the experimental setting (Section~\ref{sec:exp-setting}), we first show the benefits of GLIMPSE when labels are scarce (Section~\ref{sec:results-labels}), when observations are scarce (Section~\ref{sec:results-windows}) and when the model is small (Section~\ref{sec:results-cost}), then compare it to baselines on linear probing and full finetuning (Section~\ref{sec:results-baselines}). Finally, we ablate the two optimization paths to show the gain we get from combining them (Section~\ref{sec:results-ablation}).

\subsection{Experimental Setting}
\label{sec:exp-setting}

\paragraph{Datasets}
We evaluate on two adult cohorts that are standard benchmarks for irregularly sampled clinical time series and self-supervised EHR encoders~\citep{che2018grud,horn2020seft,tipirneni2022strats,chowdhury2023primenet,labach2023duett}, and introduce an evaluation on a paediatric cohort (Table~\ref{tab:datasets}):
\begin{itemize}
    \item \textbf{P12}~\citep{silva2012predicting}: the PhysioNet/Computing in Cardiology Challenge 2012, which provides the first 48 hours of around 12{,}000 adult ICU stays, one per patient.
    \item \textbf{MIMIC-IV}~\citep{johnson2023mimiciv}: deidentified records of adult ICU stays at the Beth Israel Deaconess Medical Center in Boston, USA. With 92{,}001 stays, it is the largest of the three cohorts.
    \item \textbf{PIC}~\citep{zeng2020pic}: paediatric and neonatal intensive care stays at the Children's Hospital of Zhejiang University School of Medicine, China, recorded between 2010 and 2018. Most of its patients are infants, and its measurement practice, reference ranges and diagnoses differ markedly from those of adult ICUs.
\end{itemize}

All three datasets are processed identically, and the model sees only time-varying physiological measurements. Demographics such as age, sex and ethnicity are deliberately excluded, as models that rely on them risk encoding and propagating existing biases in care~\citep{rajkomar2018fairness,vyas2020hidden}. Handling these biases is not in the scope of this work. Medications and treatment settings, such as mechanical ventilation, are excluded as well: they reflect decisions clinicians have already taken, and would let a model infer the outcome from the treatment instead of from the patient's condition. The physiological measurements themselves are not entirely free of such effects: which tests are ordered, and when, reflects clinical judgment, so their missingness is informative rather than random (cf. Section~\ref{sec:data}). As these measurements are the primary record of the patient's state, we retain them, and treat their missingness as part of the data rather than imputing it. To keep the reconstruction from exploiting which tests were ordered, the decoder does not distinguish masked from missing entries (Section~\ref{sec:objectives}). For P12, we use every physiological variable provided by the challenge, a compact set selected by its organizers. MIMIC-IV and PIC, in contrast, contain hundreds of laboratory tests, most of which are measured in only a handful of stays. For these two datasets, we retain the vital signs, urine output and weight, together with the 50 laboratory tests measured in the largest number of training stays, which ensures that every variable is observed often enough to be learned. Further details are provided in the Appendix Section~\ref{app:datasets}.

Each dataset is divided by patient into training (70\%), validation (15\%) and test (15\%) sets, ensuring that all stays of a patient fall into the same set. All hyperparameters, checkpoints and design choices are selected on the validation set, while the test set is reserved for the reported results.
% NOTE(selection): hier.task=stay and hier.bins=12 were tie-broken on test cells ("ahead of DuETT on every task"); on validation, s6-96 had the best mean. Either redo the tie-break on validation or state it here.
Pretraining requires no labels and therefore draws on all training stays. For the downstream tasks, in contrast, predictions are made from the first 48 hours of each stay only, following established ICU benchmarks~\citep{silva2012predicting,harutyunyan2019multitask}, and stays shorter than 48 hours are excluded (see Section \ref{sec:downstream}).

\begin{table}[t]
    \centering
    \small
    \setlength{\tabcolsep}{5pt}
    \caption{Dataset statistics. Vital: vital signs, Glasgow Coma Scale, urine output and weight. Observed entries: the fraction of the hourly time $\times$ variable grid that holds a measurement.}
    \label{tab:datasets}
    \begin{tabular}{lrrr}
        \toprule
        & P12 & MIMIC-IV & PIC \\
        \midrule
        Population & Adult & Adult & Paediatric \\
        Stays & 11{,}831 & 92{,}001 & 12{,}831 \\
        Median age (years) & 67 & 68 & 0.3 \\
        Male / female (\%) & 56 / 44 & 57 / 43 & 59 / 41 \\
        Variables (vital + lab.) & 9 + 23 & 10 + 50 & 8 + 50 \\
        Observed entries (\%) & 21.9 & 15.6 & 6.1 \\
        \midrule
        Mortality (\%) & 14.4 & 9.5 & 7.3 \\
        LOS ${\geq}7$\,d (\%) & 72.7 & 24.1 & 55.3 \\
        ICD targets & -- & 50 & 16 \\
        \bottomrule
    \end{tabular}
\end{table}

\paragraph{Downstream tasks}
We evaluate three different downstream tasks:
\begin{itemize}
    \item \textbf{Mortality}: in-hospital death, in all three datasets.
    \item \textbf{Length of stay (LOS)}: whether the ICU stay lasts at least seven days, in all three datasets. Since every evaluated stay lasts at least 48 hours, this separates stays of two to seven days from longer ones.
    \item \textbf{Diagnoses (ICD)}: a multilabel task with one binary target per three-character ICD-10 category recorded for the hospital admission, in MIMIC-IV (50 targets) and PIC (16 targets); P12 provides no diagnoses. The targets are chosen to cover a broad range of diagnoses across the main organ systems, including both acute conditions, such as sepsis or respiratory failure, and chronic comorbidities. Among these, we keep only categories with at least 50 positive stays, since rarer targets leave too few positives in the validation and test splits for a reliable estimate. Appendix~\ref{app:icd} lists all targets.
\end{itemize}

\paragraph{Protocol}
Each pretrained encoder is evaluated in two settings: linear probing, a linear head trained on the frozen representation, and full finetuning of encoder and head. All pretraining runs last 400 epochs and all finetuning runs 150 epochs, without early stopping, budgets chosen such that every method has converged by then, and the pretraining and finetuning learning rates are tuned for every method (Appendix~\ref{app:hyperparameters}). Every configuration is run with five seeds, and we report the mean and standard deviation of the test AUROC and AUPRC, macro-averaged over the labels of multilabel tasks and summarized by the mean over the eight dataset--task combinations; AUPRC results are given in Appendix~\ref{app:full-results}.

\paragraph{Baselines}
We compare against four other self-supervised encoders for ICU time series which constitute the current state-of-the-art. DuETT~\citep{labach2023duett} processes a dense variable $\times$ time-bin grid with attention alternating along the two axes and masks whole variables and time bins. AID-MAE~\citep{xiang2026aidmae} encodes only observed, unmasked values, and organizes the stay into daily records. STraTS~\citep{tipirneni2022strats} embeds every observation as a triplet of variable, time and value and is pretrained to forecast the values of the following two hours, and SMART~\citep{yu2024smart} regresses the latents of masked values, given by an EMA teacher, on an hourly grid. All four are pretrained on the same full training stays as GLIMPSE, with the same budgets, and their learning rates and masking ratios are tuned as for our models; we use their official implementations, and Appendix~\ref{app:baselines} details how we adapt them to the 48-hour window. We also report GLIMPSE-Rec, the sparse masked autoencoder on which GLIMPSE builds, trained by reconstruction alone ($\lambda_{\mathrm{pred}} = 0$). Finally, we compare with tabular models trained directly on the labels: logistic regression, XGBoost~\citep{chen2016xgboost} and three recent tabular foundation models that predict in context, TabPFN~3.5~\citep{hollmann2025tabpfn}, TabICLv2~\citep{qu2026tabiclv2} and TabFM~1.1~\citep{kong2026tabfmzeroshotfoundationmodel}. They see the same values as the encoders: the hourly grid of the first 48 hours, flattened into one row per stay with every variable in every bin as a column, and the foundation models use the entire training split as context.

\subsection{Label Efficiency}
\label{sec:results-labels}
Labelled clinical data is scarce: outcomes such as diagnoses require expert annotation, and many sites contribute only a few hundred stays. We therefore keep pretraining unchanged and restrict the labelled training stays to 1, 5, 10 and 25\%, with random subsets drawn anew for every seed, nested across fractions and shared by all methods. Figure~\ref{fig:label-efficiency} reports linear probing; full finetuning and AUPRC are given in Appendix~\ref{app:efficiency}.

Pretraining pays off most when labels are scarce: the same architecture finetuned from random weights trails fully finetuned GLIMPSE by 6.5 AUROC points with all labels and by 11.2 points with 1\% of them, and even a linear probe on GLIMPSE outperforms it by 5.1 and 11.0 points (Appendix Figure~\ref{fig:efficiency-appendix}). GLIMPSE outperforms every baseline at every fraction in both settings, and its lead over the best baseline widens as labels are removed: on linear probing the advantage goes from 0.9 AUROC points with all labels (over SMART) to 1.8 points at 1\% (over DuETT), with a peak at 5\%. With a quarter of the labels, a linear probe on GLIMPSE already matches TabPFN~3.5 trained on all of them (80.5 against 80.4), and with full finetuning it exceeds it (81.3). Over GLIMPSE-Rec, latent prediction gains more as labelled stays are added: GLIMPSE is level with it at 1\%, where a task may have a handful of positive stays, and leads it by 1.7 points with all labels.

\begin{figure*}[t]
    \centering
    \begin{subfigure}[t]{0.32\textwidth}
        \centering
        \includegraphics[width=\linewidth]{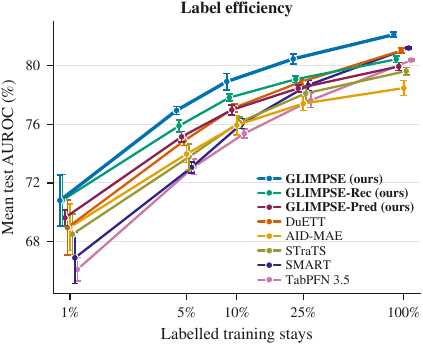}
        \caption{Label efficiency}
        \label{fig:label-efficiency}
    \end{subfigure}\hfill
    \begin{subfigure}[t]{0.32\textwidth}
        \centering
        \includegraphics[width=\linewidth]{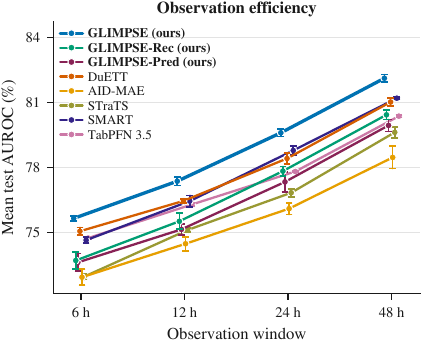}
        \caption{Observation efficiency}
        \label{fig:windows}
    \end{subfigure}\hfill
    \begin{subfigure}[t]{0.32\textwidth}
        \centering
        \includegraphics[width=\linewidth]{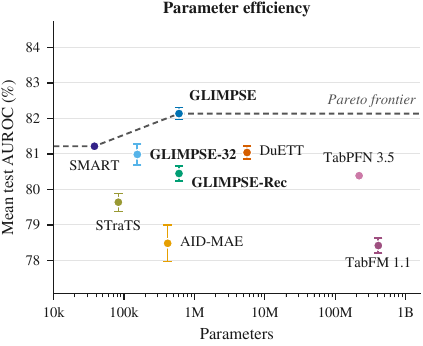}
        \caption{Parameter efficiency}
        \label{fig:cost}
    \end{subfigure}
    \caption{Mean test AUROC over the eight tasks and 5 seeds, with linear probing; full finetuning in Appendix~\ref{app:efficiency}. TabPFN~3.5 is included as the strongest tabular baseline. (a)~Trained on a fraction of the labelled training stays. (b)~Only the first 6, 12, 24 or 48 hours of each stay are observed, on the stays of at least 48 hours at every window. (c)~Against the number of parameters: the encoder for the pretrained models, the whole model for the tabular foundation models. In (a) and (b), the points of each fraction or window are shifted slightly apart for legibility.}
    \label{fig:efficiency}
\end{figure*}

\subsection{Observation Efficiency}
\label{sec:results-windows}
The earlier a prediction is made, the more useful it is, but the fewer measurements it can draw on. We therefore shorten the observation window to the first 24, 12 and 6 hours of each stay, keeping the stays of at least 48 hours so that only the input shrinks (Figure~\ref{fig:windows}; all stays that cover the window in Appendix~\ref{app:windows}).

All models lose 5.5 to 7.7 AUROC points from 48 to 6 hours, but GLIMPSE remains the best at every window in both settings. A linear probe on GLIMPSE with 6 hours of data matches GLIMPSE-Rec with 12 hours, and with 12 hours it outperforms AID-MAE with 24. TabPFN~3.5 trails GLIMPSE at every window; its gap narrows as the window shrinks, and at 6 and 12 hours it overtakes frozen GLIMPSE-Rec.

\subsection{Parameter Efficiency}
\label{sec:results-cost}
Clinical models are often deployed on local hospital infrastructure, where a small model is cheaper to train, run and audit. Figure~\ref{fig:cost} relates the mean AUROC of every model to its number of parameters. To show the performance of GLIMPSE with a smaller embedding size, we also train GLIMPSE-32, with 32- instead of 64-dimensional embeddings (154k instead of 603k parameters).

GLIMPSE lies on the Pareto frontier in both settings (full finetuning in Appendix Figure~\ref{fig:cost-full}): no model of any size reaches its mean AUROC, including DuETT with 9$\times$ and TabPFN~3.5 with 360$\times$ more parameters. GLIMPSE-32 extends the frontier to smaller models with full finetuning, where it is ahead of every baseline: with 36$\times$ fewer parameters than DuETT, it matches it with linear probing and outperforms it with full finetuning, and it outperforms AID-MAE in every task and setting except one tie. With linear probing, the smallest model, SMART (38k parameters), is slightly ahead of GLIMPSE-32 (81.2 against 81.0) and forms the small end of the frontier. The small encoder also keeps pretraining cheap: on four GPUs, GLIMPSE pretrains on MIMIC-IV in about 2.1 hours, against 3.4 to 4.3 hours for DuETT (Appendix~\ref{app:compute}).

\subsection{Comparison with baselines}
\label{sec:results-baselines}

\begin{table*}[t]
    \centering
    \footnotesize
    \setlength{\tabcolsep}{0.5pt}
    \caption{Test AUROC (\%), mean $\pm$ standard deviation over 5 seeds; Mean averages the eight tasks. Bold: best per column over all models; underline: best linear probing. AUPRC in Table~\ref{tab:main-auprc}.}
    \label{tab:main}
        \begin{tabularx}{\textwidth}{l*{9}{>{\centering\arraybackslash}X}}
        \toprule
        & \multicolumn{2}{c}{P12} & \multicolumn{3}{c}{MIMIC-IV} & \multicolumn{3}{c}{PIC} & \\
        \cmidrule(lr){2-3} \cmidrule(lr){4-6} \cmidrule(lr){7-9}
        Model & \taskicon{mortality}Mortality & \taskicon{los}LOS & \taskicon{mortality}Mortality & \taskicon{icd}ICD & \taskicon{los}LOS & \taskicon{mortality}Mortality & \taskicon{icd}ICD & \taskicon{los}LOS & Mean \\
        \midrule
        \multicolumn{10}{l}{\textit{Tabular}} \\
        LogReg & 83.4\tiny{$\pm$0.4} & 72.0\tiny{$\pm$0.4} & 83.2\tiny{$\pm$0.3} & 72.8\tiny{$\pm$0.0} & 75.7\tiny{$\pm$0.1} & 78.8\tiny{$\pm$1.4} & 81.7\tiny{$\pm$0.2} & 75.2\tiny{$\pm$0.2} & 77.8\tiny{$\pm$0.1} \\
        XGBoost & 84.2\tiny{$\pm$0.4} & 73.4\tiny{$\pm$0.7} & 85.1\tiny{$\pm$0.2} & 73.7\tiny{$\pm$0.1} & 78.8\tiny{$\pm$0.2} & 76.6\tiny{$\pm$1.5} & 79.2\tiny{$\pm$0.4} & 75.9\tiny{$\pm$0.6} & 78.4\tiny{$\pm$0.2} \\
        TabPFN 3.5 & 86.1\tiny{$\pm$0.1} & 75.7\tiny{$\pm$0.2} & 85.1\tiny{$\pm$0.2} & 75.3\tiny{$\pm$0.1} & 79.4\tiny{$\pm$0.2} & 81.0\tiny{$\pm$0.8} & 84.5\tiny{$\pm$0.3} & 76.5\tiny{$\pm$0.3} & 80.5\tiny{$\pm$0.1} \\
        TabICLv2 & 82.4\tiny{$\pm$0.3} & 70.6\tiny{$\pm$0.5} & 77.6\tiny{$\pm$1.1} & 66.2\tiny{$\pm$0.5} & 70.3\tiny{$\pm$0.4} & 76.2\tiny{$\pm$3.1} & 72.2\tiny{$\pm$1.2} & 72.0\tiny{$\pm$0.4} & 73.4\tiny{$\pm$0.8} \\
        TabFM 1.1 & 84.1\tiny{$\pm$0.2} & 73.5\tiny{$\pm$0.6} & 83.5\tiny{$\pm$0.2} & 73.2\tiny{$\pm$0.3} & 76.3\tiny{$\pm$0.4} & 81.1\tiny{$\pm$1.1} & 80.8\tiny{$\pm$0.6} & 76.2\tiny{$\pm$0.7} & 78.6\tiny{$\pm$0.1} \\
        \midrule
        \multicolumn{10}{l}{\textit{Linear probing}} \\
        DuETT & 83.5\tiny{$\pm$0.5} & 72.6\tiny{$\pm$0.3} & 86.6\tiny{$\pm$0.3} & 77.5\tiny{$\pm$0.1} & 79.9\tiny{$\pm$0.2} & 82.4\tiny{$\pm$1.3} & 88.0\tiny{$\pm$0.7} & \underline{78.8}\tiny{$\pm$0.3} & 81.1\tiny{$\pm$0.2} \\
        AID-MAE & 77.8\tiny{$\pm$1.7} & 71.4\tiny{$\pm$0.9} & 85.6\tiny{$\pm$0.5} & 75.5\tiny{$\pm$0.2} & 76.4\tiny{$\pm$0.5} & 79.2\tiny{$\pm$1.8} & 84.9\tiny{$\pm$0.7} & 77.0\tiny{$\pm$0.8} & 78.5\tiny{$\pm$0.5} \\
        STraTS & 84.7\tiny{$\pm$0.3} & 74.7\tiny{$\pm$0.7} & 86.9\tiny{$\pm$0.5} & 75.3\tiny{$\pm$0.2} & 78.5\tiny{$\pm$0.2} & 78.9\tiny{$\pm$1.1} & 81.7\tiny{$\pm$1.6} & 76.0\tiny{$\pm$0.9} & 79.6\tiny{$\pm$0.3} \\
        SMART & \underline{85.7}\tiny{$\pm$0.4} & 74.7\tiny{$\pm$0.3} & 86.4\tiny{$\pm$0.3} & 77.0\tiny{$\pm$0.2} & 79.1\tiny{$\pm$0.3} & 79.8\tiny{$\pm$1.6} & 87.2\tiny{$\pm$1.0} & 77.3\tiny{$\pm$1.1} & 80.9\tiny{$\pm$0.5} \\
        GLIMPSE-Rec (ours) & 83.4\tiny{$\pm$0.6} & 73.2\tiny{$\pm$0.5} & 86.7\tiny{$\pm$0.2} & 76.3\tiny{$\pm$0.5} & 78.2\tiny{$\pm$0.6} & 82.1\tiny{$\pm$1.0} & 86.7\tiny{$\pm$0.4} & 76.9\tiny{$\pm$0.5} & 80.4\tiny{$\pm$0.3} \\
        GLIMPSE (ours) & 85.4\tiny{$\pm$0.6} & \underline{75.0}\tiny{$\pm$0.6} & \underline{87.7}\tiny{$\pm$0.6} & \underline{77.7}\tiny{$\pm$0.3} & \underline{80.8}\tiny{$\pm$0.1} & \underline{82.8}\tiny{$\pm$1.2} & \underline{88.8}\tiny{$\pm$0.7} & 78.6\tiny{$\pm$0.3} & \underline{82.1}\tiny{$\pm$0.2} \\
        \midrule
        \multicolumn{10}{l}{\textit{Full finetuning}} \\
        DuETT & 86.3\tiny{$\pm$0.4} & 74.9\tiny{$\pm$0.5} & 88.1\tiny{$\pm$0.2} & 78.9\tiny{$\pm$0.2} & 81.8\tiny{$\pm$0.3} & 82.6\tiny{$\pm$0.6} & 88.8\tiny{$\pm$0.4} & 79.8\tiny{$\pm$0.4} & 82.6\tiny{$\pm$0.1} \\
        AID-MAE & 85.2\tiny{$\pm$0.4} & 75.9\tiny{$\pm$0.8} & 88.2\tiny{$\pm$0.2} & 78.1\tiny{$\pm$0.2} & 81.6\tiny{$\pm$0.2} & 79.1\tiny{$\pm$1.4} & 87.1\tiny{$\pm$0.6} & 78.1\tiny{$\pm$0.6} & 81.7\tiny{$\pm$0.2} \\
        STraTS & 86.1\tiny{$\pm$0.5} & 76.9\tiny{$\pm$0.5} & 87.8\tiny{$\pm$0.4} & 77.9\tiny{$\pm$0.2} & 81.2\tiny{$\pm$0.2} & 81.1\tiny{$\pm$0.7} & 88.7\tiny{$\pm$0.5} & 78.6\tiny{$\pm$0.4} & 82.3\tiny{$\pm$0.1} \\
        SMART & 86.1\tiny{$\pm$0.4} & 77.2\tiny{$\pm$0.5} & 88.1\tiny{$\pm$0.2} & 78.5\tiny{$\pm$0.1} & 81.5\tiny{$\pm$0.2} & 81.4\tiny{$\pm$1.3} & 87.2\tiny{$\pm$0.5} & 77.8\tiny{$\pm$0.4} & 82.2\tiny{$\pm$0.2} \\
        GLIMPSE-Rec (ours) & 85.9\tiny{$\pm$0.3} & 75.8\tiny{$\pm$0.1} & \textbf{89.1}\tiny{$\pm$0.3} & \textbf{79.0}\tiny{$\pm$0.1} & \textbf{82.5}\tiny{$\pm$0.3} & 82.7\tiny{$\pm$1.5} & 88.6\tiny{$\pm$0.4} & 78.8\tiny{$\pm$0.4} & 82.8\tiny{$\pm$0.2} \\
        GLIMPSE (ours) & \textbf{87.1}\tiny{$\pm$0.4} & \textbf{77.7}\tiny{$\pm$0.2} & 89.0\tiny{$\pm$0.3} & 78.8\tiny{$\pm$0.2} & 82.2\tiny{$\pm$0.3} & \textbf{84.4}\tiny{$\pm$0.9} & \textbf{89.6}\tiny{$\pm$1.0} & \textbf{80.5}\tiny{$\pm$0.7} & \textbf{83.7}\tiny{$\pm$0.2} \\
        \bottomrule
        \end{tabularx}
\end{table*}

Having seen how GLIMPSE behaves when labels, hours or parameters are limited, we now compare all methods in the standard setting, with every label and the first 48 hours of each stay. Table~\ref{tab:main} reports every task, for linear probing and full finetuning.

\paragraph{Linear Probing}
A linear head on the frozen encoder shows what pretraining alone has put into the representation. GLIMPSE gives the best representation on average and in six of the eight tasks. DuETT comes second and is slightly ahead on PIC length of stay, SMART comes third and is ahead on P12 mortality, and GLIMPSE-Rec, STraTS and AID-MAE are clearly behind; in AUPRC, GLIMPSE is also ahead of SMART and DuETT (Appendix Table~\ref{tab:main-auprc}). Adding latent prediction thus moves GLIMPSE-Rec from the lower group to the top. The comparison with SMART and DuETT is tilted against GLIMPSE: both read out one embedding per variable, so their probes see six to fifteen times as many inputs as that of GLIMPSE. Among the tabular models, TabPFN~3.5 is the strongest. On P12, whose short records are the most densely sampled (Table~\ref{tab:datasets}), it is slightly ahead of the best frozen encoders; on MIMIC-IV and PIC, it trails GLIMPSE in every task.

\paragraph{Full Finetuning}
When the encoder is finetuned together with the head, every method improves and the spread between encoders roughly halves: a finetuned encoder can learn from the labels what its pretraining missed, so the results reflect the pretraining less and the capacity to adapt more. This favours DuETT, with 9$\times$ more encoder parameters than GLIMPSE and a readout that weighs every variable separately, and halves the gap between GLIMPSE and GLIMPSE-Rec. GLIMPSE nevertheless remains the best on average in both AUROC and AUPRC, followed closely by GLIMPSE-Rec, DuETT, STraTS and SMART, with AID-MAE last, and it is the best in every task on P12 and PIC. On MIMIC-IV, with over 32{,}000 labelled training stays, GLIMPSE, GLIMPSE-Rec and DuETT are close: with this many labels, finetuning closes most of the gap that pretraining opens, the converse of the label efficiency results (Section~\ref{sec:results-labels}). Finetuned GLIMPSE also outperforms TabPFN~3.5 in every task.

\subsection{Complementary Objectives}
\label{sec:results-ablation}
GLIMPSE trains its encoder with two objectives, masked reconstruction and latent prediction. To understand what each contributes, we compare GLIMPSE with two variants that keep only one of them: GLIMPSE-Rec, trained by reconstruction alone, and GLIMPSE-Pred, trained by latent prediction alone (Section~\ref{sec:objectives}). All three share the encoder, the data and the downstream protocol, and both variants keep the pretraining settings of GLIMPSE except the learning rate, which is tuned for each of them (Appendix~\ref{app:hyperparameters}). Table~\ref{tab:ablation} reports the mean over the eight tasks, and Appendix~\ref{app:two-path} the individual tasks.

Neither objective alone matches their combination. On their own, the two objectives are close: GLIMPSE-Pred is 0.4 AUROC points below GLIMPSE-Rec with linear probing and 0.5 with full finetuning, and 0.4 points above it in AUPRC with linear probing. Combining them improves on both: GLIMPSE outperforms GLIMPSE-Rec by 1.7 AUROC points with linear probing and 0.9 with full finetuning (2.9 and 1.1 AUPRC points), and GLIMPSE-Pred by 2.1 and 1.4. The two objectives are thus complementary: each adds to the representation what the other alone does not provide.

The individual tasks suggest that the two objectives capture different aspects of a stay (Appendix~\ref{app:two-path}). With linear probing, GLIMPSE-Pred outperforms GLIMPSE-Rec on length of stay on MIMIC-IV and PIC (by 0.9 and 1.6 points), a property of the overall course of the stay, but trails it by 0.5 to 2.0 points on mortality and diagnoses, which may depend more on the details of individual measurements that reconstruction is trained to retain. On P12, whose records end after 48 hours, it is level on mortality and behind on length of stay.
% We believe that reconstruction keeps the representation tied to the values of individual measurements, on which many outcomes directly depend: acute kidney injury, for instance, is diagnosed from the rise in serum creatinine. Latent prediction, in contrast, encourages a summary of the patient's overall physiological state and its trajectory, which no single measurement captures. GLIMPSE benefits from both, and its representation is better than either objective provides alone.

\begin{table}[t]
    \centering
    \footnotesize
    \setlength{\tabcolsep}{3pt}
    \caption{Ablation of the two pretraining objectives: mean test AUROC and AUPRC (\%) over the eight tasks, mean $\pm$ standard deviation over 5 seeds. Rec.: masked reconstruction; Pred.: latent prediction.}
    \label{tab:ablation}
    \resizebox{\columnwidth}{!}{%
    \begin{tabular}{lcccccc}
        \toprule
        & & & \multicolumn{2}{c}{Linear probing} & \multicolumn{2}{c}{Full finetuning} \\
        \cmidrule(lr){4-5} \cmidrule(lr){6-7}
        Variant & Rec. & Pred. & AUROC & AUPRC & AUROC & AUPRC \\
        \midrule
        GLIMPSE & \checkmark & \checkmark & \textbf{82.1}\tiny{$\pm$0.2} & \textbf{52.1}\tiny{$\pm$0.3} & \textbf{83.7}\tiny{$\pm$0.2} & \textbf{54.6}\tiny{$\pm$0.3} \\
        GLIMPSE-Rec & \checkmark & -- & 80.4\tiny{$\pm$0.3} & 49.2\tiny{$\pm$0.6} & 82.8\tiny{$\pm$0.2} & 53.5\tiny{$\pm$0.4} \\
        GLIMPSE-Pred & -- & \checkmark & 80.0\tiny{$\pm$0.3} & 49.6\tiny{$\pm$1.1} & 82.3\tiny{$\pm$0.3} & 53.3\tiny{$\pm$0.2} \\
        \bottomrule
    \end{tabular}}
\end{table}

\section{Discussion}
\label{sec:discussion}

The advantages of GLIMPSE are largest where the downstream task offers the least supervision: with a frozen encoder, few labels, short observation windows and on the smaller cohorts, P12 and PIC. There, the representation must already carry what the task needs, and training the representation of a few hours to agree with that of the whole stay makes early, sparsely labelled predictions easier. On PIC, at a fixed specificity of 90\%, finetuned GLIMPSE identifies 4 and 5 more of the 73 deaths in the test set than DuETT and SMART respectively, the strongest baselines.

GLIMPSE's objective puts emphasis on encoding information of the whole stay, including informative measurements near the end of the stay. However, as can be seen for the P12 results, it performs better even when the availability of whole stay information is limited to the first 48 hours for pretraining as well. This suggests that the benefit comes from predicting representations of other parts of the stay in general, not of the whole stay specifically: predicting the next window or the end of the stay instead performs similarly, although the whole stay works best (Appendix~\ref{app:ablations}). When labels are abundant and the encoder is finetuned, as on MIMIC-IV, the methods converge.

\paragraph{Limitations and future work.}
The downstream tasks are classifications from the first hours of stays of at least 48 hours; time-to-event outcomes and predictions updated throughout the stay are not covered, although its accuracy from a few hours makes GLIMPSE a natural candidate for continuous risk monitoring. The input is restricted to physiological measurements. Excluding demographics and treatments removes two sources of bias and leakage, but the timing of measurements still reflects clinical practice, and performance across patient subgroups, which must be assessed before any clinical use, was not evaluated. Future work will explore pretraining across institutions and transfer between them, to assess whether one model can serve both adult and paediatric care, as well as a wider range of prediction targets and context scales.

\section{Conclusion}
\label{sec:conclusion}

We introduced GLIMPSE, a sparse encoder for intensive-care records that, besides reconstructing individual measurements, infers the representation of the whole stay from a glimpse of it. On two adult and one paediatric cohort, it performs best on average when representations are probed and after full finetuning, and its advantage holds where labels, observation time and parameters are scarce. Predicting latent representations thus complements reconstruction: the two objectives together learn better clinical representations than either does alone.

\section*{Software and Data}
The code will be made publicly available upon publication.

% Remove the acknowledgements in the blind submission version.
\section*{Acknowledgements}
We thank the creators of MIMIC-IV \citep{johnson2023mimiciv,johnson2024mimicivdata}, the Paediatric Intensive Care database \citep{zeng2020pic,li2020picdata} and the PhysioNet/Computing in Cardiology Challenge 2012 \citep{silva2012predicting} for making these data available through PhysioNet \citep{pollard2026physionet}. This work was supported by computational resources provided by the Swiss National Supercomputing Centre (CSCS) under the Alps infrastructure through the Swiss AI Initiative (project ID aa012).

\section*{Impact Statement}
This work develops representation learning for intensive-care records, with the aim of supporting clinical risk prediction where labelled data, observation time or computing resources are limited, including in paediatric care, which is underrepresented in the evaluation of EHR models. The models are research artefacts and are not intended for clinical decision-making without prospective validation, calibration and regulatory review. We deliberately exclude demographics and treatments from the input, so that the models do not condition on a demographic profile or infer the outcome from clinical decisions, but measurement patterns still reflect site-specific practice and may encode disparities in care; performance across patient subgroups has not been assessed and must be before any deployment. All three datasets are deidentified and accessed under their data use agreements, and no patient data are redistributed with this work.

\bibliography{references}
\bibliographystyle{icml2026}

\newpage
\appendix
\onecolumn
% Appendix structure, following the main text:
% A Datasets what the models see and predict (Section 4.1)
% B GLIMPSE and GLIMPSE-Rec settings, hyperparameter sweeps, design of the objective, readout, compute
% C Baselines how every baseline is run, and why it is a fair comparison
% D Extended results everything behind the figures and tables of Section 4, in the order of Section 4

\section{Datasets}
\label{app:datasets}

We use ICU stays from three datasets that differ in their patient populations and in the measurements they record (Table~\ref{tab:datasets}). Each stay is one example. The model sees only the time-varying physiological measurements and the times at which they were taken. We leave out static descriptors such as age and sex, as well as medications and treatment settings such as ventilation, so that the representation is learned from the patient's measured physiological state rather than from a demographic profile or a record of interventions.

\subsection{Preprocessing}
\label{app:preprocessing}

\paragraph{Shared temporal processing} The prepared data keep the source timestamps, and the measurements are grouped into hourly bins, counted from ICU admission, when an experiment loads them. Within a bin we keep the latest observed value of each measurement, following established ICU benchmarks~\citep{harutyunyan2019multitask}; urine output is the exception and is summed, since it records an accumulated volume. Measurements taken outside the ICU stay are dropped, and bins without any measurement are omitted. Missing values remain missing, without forward filling or interpolation. During data preparation, clearly implausible values, such as a pH of $2.2 \times 10^{10}$ in PIC, are removed with fixed bounds that are wide enough to retain clinically extreme observations. Each channel is then clipped at its 5th and 95th percentiles and standardized with its mean and standard deviation after clipping. These statistics are computed on the training stays of each dataset and applied unchanged to validation and test data.

\paragraph{Cohort and input window} Pretraining uses all training stays, whatever their length, and may use the whole stay. The downstream tasks use only the stays of at least 48 hours, in every split, and only their first 48 hours, so that the end of the input cannot reveal the length of stay. On P12 the two coincide, because the challenge records end after 48 hours.

\subsection{Data splits}
\label{app:data-splits}
We split each dataset by patient: 70\% of the patients are assigned to training, 15\% to validation and 15\% to test. All stays of a patient, including readmissions, therefore fall in the same split. Because the 48-hour filter is applied after splitting, the stays of at least 48 hours do not follow the 70/15/15 proportions exactly: 8{,}277 / 1{,}774 / 1{,}775 training, validation and test stays on P12, 4{,}903 / 1{,}039 / 1{,}044 on PIC and 32{,}228 / 6{,}902 / 7{,}073 on MIMIC-IV. The laboratory measurements are selected on the training split before the filter, and the normalization statistics on all training stays. The validation split is used to select checkpoints and hyperparameters, and the test split only for the reported results.

\subsection{MIMIC-IV-ICU\texorpdfstring{~\citep{johnson2023mimiciv}}{}}
MIMIC-IV contains deidentified hospital and intensive care records from the Beth Israel Deaconess Medical Center in Boston. We use its adult ICU stays: 92{,}001 in total, of which 46{,}203 last at least 48 hours.

\paragraph{Input variables} The input has 60 channels: seven vital signs (heart rate, respiratory rate, systolic, diastolic and mean blood pressure, peripheral oxygen saturation and temperature), the Glasgow Coma Scale, urine output, weight and 50 laboratory measurements. To select the laboratory measurements, we map the source item identifiers to measurement concepts, keeping assays of different fluids separate (for example, urine pH and blood pH, or whole-blood and serum sodium). Specimen-interference indices, reference values, therapeutic drug concentrations, inspired oxygen fraction and ventilator settings are excluded. We then rank the remaining concepts by the number of training stays in which they are observed and keep the 50 most prevalent.
Table~\ref{tab:labs} lists the selected laboratory measurements of every dataset.

\begin{table}[htbp]
    \centering
    \footnotesize
    \caption{Laboratory measurements of each dataset. MIMIC-IV and PIC: the 50 measurements observed in the most training stays, in decreasing order; assays of different fluids are separate measurements. P12: all laboratory measurements of the challenge.}
    \label{tab:labs}
    \begin{tabularx}{\textwidth}{@{}lX@{}}
        \toprule
        Dataset & Laboratory measurements \\
        \midrule
        P12 (23) & Albumin, Alkaline Phosphatase, Alanine Aminotransferase (ALT), Aspartate Aminotransferase (AST), Bicarbonate, Bilirubin (total), Cholesterol, Creatinine, Glucose, Hematocrit, Lactate, Magnesium, PaCO$_2$, PaO$_2$, pH, Platelet Count, Potassium, SaO$_2$, Sodium, Troponin I, Troponin T, Urea Nitrogen, White Blood Cells. \\
        \midrule
        MIMIC-IV (50) & Glucose, Sodium, Potassium, Chloride, Creatinine, Bicarbonate, Urea Nitrogen, Anion Gap, Hematocrit, Hemoglobin, Platelet Count, WBC Count, Red Blood Cells, MCV, MCHC, MCH, RDW, Magnesium, Phosphate, Calcium (total), PT, INR(PT), PTT, pH, Lactate, pO$_2$, Base Excess, pCO$_2$, Calculated Total CO$_2$, RDW-SD, Aspartate Aminotransferase (AST), Alanine Aminotransferase (ALT), Bilirubin (total), Alkaline Phosphatase, Ionized Calcium, Lymphocytes (\%), Monocytes (\%), Basophils (\%), Eosinophils (\%), Neutrophils (\%), Lactate Dehydrogenase (LD), Albumin, pH (urine), Specific Gravity (urine), Potassium (whole blood), Oxygen Saturation, Creatine Kinase (CK), Creatine Kinase MB Isoenzyme (CK-MB), Fibrinogen (functional), Sodium (whole blood). \\
        \midrule
        PIC (50) & Hematocrit, Hemoglobin, Base Excess, pH, pCO$_2$, Chloride (whole blood), Sodium (whole blood), Potassium (whole blood), Ionized Calcium, Calculated Bicarbonate (whole blood), Carboxyhemoglobin, Oxygen Saturation, Methemoglobin, pO$_2$, Bicarbonate, Lactate, Anion Gap, Red Blood Cells, Platelet Count, WBC Count, MCH, MCHC, MCV, RDW, Glucose, Neutrophils (\%), Basophils (\%), Lymphocytes (\%), Eosinophils (\%), Monocytes (\%), Plateletcrit (PCT), Platelet Distribution Width, Basophil Count, Eosinophil Count, Monocyte Count, Lymphocyte Count, Neutrophil Count, Mean Platelet Volume (MPV), C-Reactive Protein, Alanine Aminotransferase (ALT), Urea, Creatinine, Bilirubin (total), INR(PT), PT, Protein (total), Aspartate Aminotransferase (AST), Prealbumin, Albumin/Globulin Ratio, Albumin. \\
        \bottomrule
    \end{tabularx}
\end{table}

\subsection{Paediatric Intensive Care (PIC)\texorpdfstring{~\citep{zeng2020pic}}{}}
PIC contains records from paediatric and neonatal intensive care at the Children's Hospital of Zhejiang University School of Medicine in China, collected between 2010 and 2018. Most of its patients are infants. The cohort contains 12{,}831 stays, of which 6{,}986 last at least 48 hours.

\paragraph{Input variables} The input has 58 channels: six vital signs (those of MIMIC-IV-ICU except mean blood pressure), urine output, weight and 50 laboratory measurements, selected in the same way as for MIMIC-IV-ICU. PIC does not record the Glasgow Coma Scale. Haemoglobin and mean corpuscular haemoglobin concentration reported in g/L are converted to g/dL.

\subsection{PhysioNet 2012 Challenge (P12)\texorpdfstring{~\citep{silva2012predicting}}{}}
The challenge provides the first 48 hours of adult ICU stays that lasted at least 48 hours. We use 11{,}831 of its 12{,}000 records, of which 11{,}826 satisfy our duration criterion, and we split them ourselves instead of using the challenge's sets A, B and C. Each patient has a single record.

\paragraph{Input variables} The input has 32 channels: six vital signs (heart rate, respiratory rate, systolic, diastolic and mean blood pressure, and temperature), the Glasgow Coma Scale, urine output, weight and 23 laboratory measurements. Invasive and noninvasive blood pressure are combined into one channel each, preferring the invasive measurement when both are recorded at the same time. We exclude inspired oxygen fraction and the mechanical-ventilation indicator as treatment variables, and age, sex, height and ICU type as static descriptors.

\subsection{Tasks and labels}
\label{app:icd}

\paragraph{Mortality and length of stay} In-hospital mortality is a binary label in every dataset, and so is length of stay: whether the stay lasts at least seven days. Since the evaluated stays last at least 48 hours, its negative class covers stays of two to seven days. As for every ICD target, we report the AUROC and the AUPRC of the positive class. The models predict length of stay as one categorical output over its classes, which at the windows below 48 hours of Appendix~\ref{app:windows} include a third class, stays shorter than two days; there, the AUROC is macro-averaged over the classes and the AUPRC is that of stays of at least seven days. The AUPRC of length of stay follows its prevalence (Table~\ref{tab:datasets}): stays of at least seven days are the majority on P12 and PIC and the minority on MIMIC-IV.

\paragraph{ICD labels} ICD classification is a multilabel task in MIMIC-IV-ICU and PIC, with one binary target per three-character ICD-10 category; P12 has no ICD task. We use the categories rather than the full codes because the full codes split a condition into many subcodes, for example sepsis (\texttt{A41}) by causative organism, and most subcodes have too few positive stays to train a classifier and to evaluate it reliably, in particular in the validation and test splits. A target is positive when the stay's diagnosis list contains a code of that category, and several targets may be positive for the same stay. The diagnoses are recorded for the whole hospital admission, so they are not limited to conditions that arise within the 48-hour input window. A stay without a diagnosis list (14 in MIMIC-IV-ICU and 5 in PIC) has a missing label for this task but is kept for the other tasks, whereas a stay whose list contains none of the targets is negative for all of them.

\paragraph{PIC targets} PIC has its own set of 16 targets from nine chapter letters. Half of them are perinatal or congenital conditions; the others include pneumonia, sepsis, encephalitis and convulsions. We chose them to cover a wide range of diseases and organ systems, among the most frequent categories in the training stays of at least 48 hours so that every target has enough positive examples in the validation and test splits. The rarest target, neonatal bacterial sepsis (\texttt{P36}), occurs in 50 stays, and the most frequent, disorders related to short gestation and low birth weight (\texttt{P07}), in 947 (Table~\ref{tab:pic-icd3}). PIC uses the Chinese extension of ICD-10, whose codes have six characters (for example \texttt{J18900}); as for MIMIC-IV-ICU, the category is given by the first three.

\paragraph{MIMIC-IV-ICU targets} The diagnosis task has 50 binary targets from 13 ICD-10 chapter letters, chosen to cover a wide range of diseases and organ systems: acute conditions such as sepsis, acute kidney failure and respiratory failure as well as chronic comorbidities such as hypertension, diabetes and chronic kidney disease. The categories were chosen by clinical judgement and fixed before any model was trained. The rarest target occurs in 131 stays of at least 48 hours. Table~\ref{tab:mimic-icd3} lists every target with its frequency. The diagnosis lists contain not only the codes themselves but also their ancestors in the ICD-10 hierarchy, including blocks such as \texttt{E86--E87}; we skip the blocks, since they are not categories.

\begin{table}[t]
    \centering
    \footnotesize
    \caption{PIC ICD targets, grouped by ICD-10 chapter: number of stays with the target in the training, validation and test splits, and in total, with the percentage of the 6{,}981 stays of at least 48 hours with a diagnosis list. A stay may have several targets.}
    \label{tab:pic-icd3}
    \begin{tabularx}{\textwidth}{@{}l l X r r r r@{}}
        \toprule
        Chapter & Code & Category & Train & Val. & Test & Total (\%) \\
        \midrule
        A\enspace Infectious diseases & \texttt{A41} & Other sepsis & 92 & 19 & 17 & 128 (1.8) \\
        \addlinespace[3pt]
        C\enspace Neoplasms & \texttt{C95} & Leukemia of unspecified cell type & 83 & 11 & 24 & 118 (1.7) \\
        \addlinespace[3pt]
        D\enspace Blood and immune & \texttt{D64} & Other anemias & 38 & 8 & 6 & 52 (0.7) \\
        \addlinespace[3pt]
        G\enspace Nervous system & \texttt{G04} & Encephalitis, myelitis and encephalomyelitis & 107 & 33 & 26 & 166 (2.4) \\
        \addlinespace[3pt]
        \multirow[t]{2}{*}{I\enspace Circulatory system} & \texttt{I51} & Complications and ill-defined descriptions of heart disease & 145 & 23 & 33 & 201 (2.9) \\
         & \texttt{I62} & Other nontraumatic intracranial hemorrhage & 35 & 7 & 9 & 51 (0.7) \\
        \addlinespace[3pt]
        J\enspace Respiratory system & \texttt{J18} & Pneumonia, organism unspecified & 515 & 116 & 123 & 754 (10.8) \\
        \addlinespace[3pt]
        \multirow[t]{5}{*}{P\enspace Perinatal conditions} & \texttt{P07} & Disorders related to short gestation and low birth weight & 660 & 139 & 148 & 947 (13.6) \\
         & \texttt{P21} & Birth asphyxia & 118 & 16 & 21 & 155 (2.2) \\
         & \texttt{P22} & Respiratory distress of newborn & 91 & 14 & 12 & 117 (1.7) \\
         & \texttt{P36} & Bacterial sepsis of newborn & 30 & 9 & 11 & 50 (0.7) \\
         & \texttt{P77} & Necrotizing enterocolitis of fetus and newborn & 45 & 15 & 10 & 70 (1.0) \\
        \addlinespace[3pt]
        \multirow[t]{3}{*}{Q\enspace Congenital malformations} & \texttt{Q21} & Congenital malformations of cardiac septa & 326 & 96 & 71 & 493 (7.1) \\
         & \texttt{Q24} & Other congenital malformations of heart & 272 & 53 & 46 & 371 (5.3) \\
         & \texttt{Q43} & Other congenital malformations of intestine & 38 & 9 & 10 & 57 (0.8) \\
        \addlinespace[3pt]
        R\enspace Symptoms and signs & \texttt{R56} & Convulsions, not elsewhere classified & 119 & 27 & 41 & 187 (2.7) \\
        \bottomrule
    \end{tabularx}
\end{table}

\begin{table}[p]
    \centering
    \footnotesize
    \caption{MIMIC-IV-ICU ICD targets, as in Table~\ref{tab:pic-icd3}; percentages of the 46{,}189 stays of at least 48 hours with a diagnosis list.}
    \label{tab:mimic-icd3}
    \begin{tabularx}{\textwidth}{@{}l l X r r r r@{}}
        \toprule
        Chapter & Code & Category & Train & Val. & Test & Total (\%) \\
        \midrule
        A\enspace Infectious diseases & \texttt{A41} & Other sepsis & 7{,}221 & 1{,}532 & 1{,}558 & 10{,}311 (22.3) \\
        \addlinespace[3pt]
        \multirow[t]{2}{*}{C\enspace Neoplasms} & \texttt{C34} & Malignant neoplasm of bronchus and lung & 799 & 190 & 203 & 1{,}192 (2.6) \\
         & \texttt{C78} & Secondary malignancy of respiratory and digestive organs & 1{,}116 & 282 & 246 & 1{,}644 (3.6) \\
        \addlinespace[3pt]
        \multirow[t]{3}{*}{D\enspace Blood and immune} & \texttt{D50} & Iron deficiency anemia & 1{,}634 & 331 & 376 & 2{,}341 (5.1) \\
         & \texttt{D62} & Acute posthemorrhagic anemia & 7{,}382 & 1{,}554 & 1{,}667 & 10{,}603 (23.0) \\
         & \texttt{D64} & Other anemias & 4{,}914 & 1{,}108 & 1{,}136 & 7{,}158 (15.5) \\
        \addlinespace[3pt]
        \multirow[t]{6}{*}{E\enspace Endocrine and metabolic} & \texttt{E03} & Other hypothyroidism & 4{,}291 & 918 & 945 & 6{,}154 (13.3) \\
         & \texttt{E11} & Type 2 diabetes mellitus & 9{,}705 & 2{,}141 & 2{,}039 & 13{,}885 (30.1) \\
         & \texttt{E66} & Overweight and obesity & 4{,}138 & 937 & 890 & 5{,}965 (12.9) \\
         & \texttt{E78} & Disorders of lipoprotein metabolism and other lipidemias & 13{,}896 & 2{,}939 & 3{,}021 & 19{,}856 (43.0) \\
         & \texttt{E86} & Volume depletion & 2{,}684 & 576 & 574 & 3{,}834 (8.3) \\
         & \texttt{E87} & Other fluid, electrolyte and acid-base disorders & 15{,}681 & 3{,}407 & 3{,}530 & 22{,}618 (49.0) \\
        \addlinespace[3pt]
        \multirow[t]{4}{*}{F\enspace Mental and behavioural} & \texttt{F03} & Unspecified dementia & 911 & 178 & 196 & 1{,}285 (2.8) \\
         & \texttt{F05} & Delirium due to a known physiological condition & 2{,}617 & 546 & 620 & 3{,}783 (8.2) \\
         & \texttt{F10} & Alcohol-related disorders & 3{,}193 & 711 & 720 & 4{,}624 (10.0) \\
         & \texttt{F32} & Depressive episode & 4{,}979 & 1{,}029 & 1{,}093 & 7{,}101 (15.4) \\
        \addlinespace[3pt]
        \multirow[t]{4}{*}{G\enspace Nervous system} & \texttt{G20} & Parkinson disease & 374 & 70 & 74 & 518 (1.1) \\
         & \texttt{G40} & Epilepsy and recurrent seizures & 2{,}066 & 468 & 456 & 2{,}990 (6.5) \\
         & \texttt{G45} & Transient cerebral ischemic attacks and related syndromes & 96 & 17 & 18 & 131 (0.3) \\
         & \texttt{G93} & Other disorders of brain & 5{,}647 & 1{,}146 & 1{,}196 & 7{,}989 (17.3) \\
        \addlinespace[3pt]
        \multirow[t]{8}{*}{I\enspace Circulatory system} & \texttt{I10} & Essential (primary) hypertension & 12{,}801 & 2{,}651 & 2{,}795 & 18{,}247 (39.5) \\
         & \texttt{I21} & Acute myocardial infarction & 3{,}804 & 728 & 833 & 5{,}365 (11.6) \\
         & \texttt{I25} & Chronic ischemic heart disease & 10{,}186 & 2{,}143 & 2{,}243 & 14{,}572 (31.5) \\
         & \texttt{I26} & Pulmonary embolism & 1{,}101 & 225 & 260 & 1{,}586 (3.4) \\
         & \texttt{I48} & Atrial fibrillation and flutter & 11{,}030 & 2{,}342 & 2{,}413 & 15{,}785 (34.2) \\
         & \texttt{I50} & Heart failure & 10{,}277 & 2{,}257 & 2{,}256 & 14{,}790 (32.0) \\
         & \texttt{I63} & Cerebral infarction & 2{,}414 & 504 & 488 & 3{,}406 (7.4) \\
         & \texttt{I82} & Other venous embolism and thrombosis & 2{,}058 & 483 & 475 & 3{,}016 (6.5) \\
        \addlinespace[3pt]
        \multirow[t]{7}{*}{J\enspace Respiratory system} & \texttt{J18} & Pneumonia, organism unspecified & 4{,}401 & 951 & 892 & 6{,}244 (13.5) \\
         & \texttt{J44} & Other chronic obstructive pulmonary disease & 4{,}901 & 1{,}094 & 1{,}068 & 7{,}063 (15.3) \\
         & \texttt{J45} & Asthma & 2{,}176 & 527 & 424 & 3{,}127 (6.8) \\
         & \texttt{J69} & Pneumonitis due to solids and liquids & 3{,}473 & 759 & 799 & 5{,}031 (10.9) \\
         & \texttt{J80} & Acute respiratory distress syndrome & 688 & 161 & 136 & 985 (2.1) \\
         & \texttt{J81} & Pulmonary edema & 1{,}485 & 321 & 330 & 2{,}136 (4.6) \\
         & \texttt{J96} & Respiratory failure, not elsewhere classified & 12{,}205 & 2{,}762 & 2{,}705 & 17{,}672 (38.3) \\
        \addlinespace[3pt]
        \multirow[t]{7}{*}{K\enspace Digestive system} & \texttt{K21} & Gastroesophageal reflux disease & 6{,}663 & 1{,}423 & 1{,}472 & 9{,}558 (20.7) \\
         & \texttt{K52} & Other noninfective gastroenteritis and colitis & 429 & 102 & 115 & 646 (1.4) \\
         & \texttt{K56} & Paralytic ileus and intestinal obstruction without hernia & 1{,}921 & 398 & 421 & 2{,}740 (5.9) \\
         & \texttt{K70} & Alcoholic liver disease & 1{,}573 & 346 & 356 & 2{,}275 (4.9) \\
         & \texttt{K72} & Hepatic failure, not elsewhere classified & 1{,}851 & 391 & 415 & 2{,}657 (5.8) \\
         & \texttt{K74} & Fibrosis and cirrhosis of liver & 1{,}234 & 274 & 286 & 1{,}794 (3.9) \\
         & \texttt{K92} & Other diseases of digestive system & 1{,}879 & 410 & 416 & 2{,}705 (5.9) \\
        \addlinespace[3pt]
        \multirow[t]{3}{*}{N\enspace Genitourinary system} & \texttt{N17} & Acute kidney failure & 12{,}685 & 2{,}790 & 2{,}677 & 18{,}152 (39.3) \\
         & \texttt{N18} & Chronic kidney disease & 7{,}654 & 1{,}691 & 1{,}707 & 11{,}052 (23.9) \\
         & \texttt{N39} & Other disorders of urinary system & 5{,}287 & 1{,}132 & 1{,}173 & 7{,}592 (16.4) \\
        \addlinespace[3pt]
        \multirow[t]{2}{*}{R\enspace Symptoms and signs} & \texttt{R57} & Shock, not elsewhere classified & 4{,}319 & 947 & 959 & 6{,}225 (13.5) \\
         & \texttt{R65} & Symptoms and signs of systemic inflammation and infection & 6{,}259 & 1{,}354 & 1{,}313 & 8{,}926 (19.3) \\
        \addlinespace[3pt]
        \multirow[t]{2}{*}{S\enspace Injury} & \texttt{S06} & Intracranial injury & 1{,}515 & 320 & 305 & 2{,}140 (4.6) \\
         & \texttt{S72} & Fracture of femur & 357 & 68 & 75 & 500 (1.1) \\
        \addlinespace[3pt]
        T\enspace Complications of care & \texttt{T81} & Complications of procedures, not elsewhere classified & 2{,}171 & 410 & 448 & 3{,}029 (6.6) \\
        \bottomrule
    \end{tabularx}
\end{table}

\clearpage
\section{Implementation of GLIMPSE and GLIMPSE-Rec}
\label{app:implementation}

\subsection{Method details}
\label{app:method-details}
\paragraph{Value normalization.} We clip each value $x_{b,d}$ at the 5th and 95th percentiles $q^{5}_d$ and $q^{95}_d$ of its variable~\citep{xiang2026aidmae} and then standardize it,
\begin{equation*}
    z_{b,d} = \frac{\min\!\big(\max(x_{b,d},\, q^{5}_d),\, q^{95}_d\big) - \mu_d}{\sigma_d},
\end{equation*}
where $\mu_d$ and $\sigma_d$ are the mean and standard deviation of the clipped values. All statistics are computed on the training stays of each dataset.

\paragraph{Time embedding.} The time $\tau$ of a bin, in hours since admission, is embedded with $E/2$ sine and $E/2$ cosine functions~\citep{vaswani2017attention} with periods spaced logarithmically from $P_{\min}=2$ hours to $P_{\max}=30$ days,
\begin{equation*}
    \gamma(\tau) = \big[\sin(2\pi\tau/P_k)\big]_{k=1}^{E/2} \,\Vert\, \big[\cos(2\pi\tau/P_k)\big]_{k=1}^{E/2},
    \qquad
    P_k = P_{\min} \left(\frac{P_{\max}}{P_{\min}}\right)^{\frac{k-1}{E/2-1}},
\end{equation*}
where $\Vert$ denotes concatenation. The short periods resolve changes within hours, the long ones the course of a stay over days to weeks.

\paragraph{Windows of the latent prediction.} Every window is divided into $K=12$ bins, whatever its duration: a 12-hour context at hourly resolution, a 24-hour context at two-hour and a 48-hour context at four-hour resolution, and the whole stay at the smallest power-of-two number of hours per bin that covers it. The cost of a window thus does not grow with its duration. Within a coarser bin, values are aggregated as for the hourly bins (latest value, volumes summed) and normalized as above, and each bin keeps its time since admission. Which single context length works best depends on the dataset: 12 hours on P12 and 48 hours on MIMIC-IV and PIC (Appendix~\ref{app:ablations}). Rather than tuning this length for every dataset, we sample it between 12 and 48 hours, which covers the length of the downstream window. This makes the objective robust to the choice and performs within noise to the best single length on average over the datasets. The prediction is made in representation space rather than in the space of values, so the model need not forecast every future measurement, which is largely unpredictable, but only a summary of the stay.

\paragraph{Loss of the latent prediction.} As in BYOL~\citep{grill2020byol}, we use the cosine distance rather than a squared error: it compares only the directions of the two representations, so the loss cannot be lowered by shrinking their norms, and it is insensitive to differences in scale between the encoder and its slowly moving average.

\subsection{Architecture and training}
Table~\ref{tab:hyperparameters} lists the settings of GLIMPSE and GLIMPSE-Rec. GLIMPSE-Rec is GLIMPSE with $\lambda_{\mathrm{pred}} = 0$, without the predictor and the target encoder; everything else is shared, except the pretraining learning rate, which is selected for each, and the pretraining checkpoint. GLIMPSE-Rec keeps the checkpoint with the lowest validation reconstruction loss. GLIMPSE keeps the last one, because the validation loss of latent prediction behaves differently: it typically drops quickly early in training, while the representation is not yet useful, then rises again and finally decreases slightly or levels off, as for GLIMPSE on MIMIC-IV. A low validation loss of a latent prediction objective therefore does not indicate a good representation~\citep{thilak2024lidar,ennadir2026early}.

\begin{table}[htbp]
    \centering
    \small
    \caption{Settings of GLIMPSE and GLIMPSE-Rec. Batch size, optimizer, weight decay and gradient clipping are the same for downstream training.}
    \label{tab:hyperparameters}
    \begin{tabular}{ll}
        \toprule
        Setting & Value \\
        \midrule
        \multicolumn{2}{l}{\textit{Encoder and decoder}} \\
        Embedding dimension $E$ & 64 \\
        Encoder & 12 layers, 4 heads \\
        Decoder & 4 layers, 4 heads \\
        Decoder input & every variable in every bin of the sample \\
        Readout & \texttt{[CLS]} and mean of the value tokens (Eq.~\ref{eq:readout}) \\
        \midrule
        \multicolumn{2}{l}{\textit{Pretraining}} \\
        Masking ratio & 0.75 of the observed entries \\
        Reconstruction sample & at most 48 hourly bins of the stay \\
        Epochs & 400, no early stopping \\
        Batch size & 128 \\
        Optimizer & AdamW, $\beta = (0.9, 0.95)$, weight decay 0.05 \\
        Learning rate & $3 \cdot 10^{-2}$ (GLIMPSE), $10^{-3}$ (GLIMPSE-Rec) \\
        Schedule & 20 warmup epochs, cosine decay to 0 \\
        Gradient clipping & norm 1.0 \\
        Checkpoint & last (GLIMPSE), lowest validation reconstruction loss (GLIMPSE-Rec) \\
        \midrule
        \multicolumn{2}{l}{\textit{Latent prediction (GLIMPSE)}} \\
        Weight $\lambda_{\mathrm{pred}}$ & 0.3 \\
        Context lengths $s$ & 12, 24, 48 hours \\
        Bins per window $K$ & 12 \\
        Target & whole stay, target encoder \\
        Target encoder momentum $m$ & 0.999 \\
        Predictor $g$ & MLP, $E \to 4E \to E$, GELU \\
        \midrule
        \multicolumn{2}{l}{\textit{Downstream}} \\
        Input & first 48 hours, all observed values, no masking \\
        Epochs & 150, no early stopping \\
        Head learning rate & $10^{-2}$ (linear probing and full finetuning) \\
        Encoder learning rate & $0.01 \times$ head learning rate (full finetuning) \\
        Encoder dropout & 0.1 (full finetuning) \\
        Schedule & 5 warmup epochs, cosine decay to 0 \\
        Checkpoint & highest validation AUROC \\
        \bottomrule
    \end{tabular}
\end{table}

\subsection{Hyperparameter Selection}
\label{app:hyperparameters}
We tune, for every self-supervised encoder, the pretraining learning rate and the head learning rate of full finetuning, and for GLIMPSE additionally the weight $\lambda_{\mathrm{pred}}$ of the prediction loss, jointly with the learning rate. For GLIMPSE, GLIMPSE-Rec, AID-MAE and SMART we also tune the masking ratio. Every choice maximizes the mean validation AUROC over the eight dataset--task combinations. Pretraining choices are scored by a shared linear probe (logistic regression) on the frozen representation, which is fast and identical for all methods; for the finetuning learning rate, we take the smallest rate within noise.

The diagnosis task of PIC first had 50 targets. Many of them had very few positive stays in the validation and test splits, which made their AUROC noisy, so we reduced them to the 16 targets of Table~\ref{tab:pic-icd3}. The learning rate sweeps of GLIMPSE, GLIMPSE-Rec, DuETT and AID-MAE (Tables~\ref{tab:lr} and~\ref{tab:weight-lr}) and the masking sweep of AID-MAE were run on the earlier 50 targets and not repeated because of limited compute resources. STraTS and SMART were added later and tuned on the 16 targets, so their rows are listed separately in these tables; GLIMPSE-Pred was also tuned on the 16 targets. All other results use the 16 targets.

Table~\ref{tab:lr} gives the learning rate sweeps. Every selected pretraining rate is surrounded by worse rates on both sides, except for DuETT, whose two best pretraining rates tie ($10^{-3}$ and $3{\cdot}10^{-3}$, 80.95 both). STraTS~\citep{tipirneni2022strats} prefers $3{\cdot}10^{-3}$, six times its original rate, and SMART~\citep{yu2024smart} is insensitive to the pretraining rate, within 0.2 points from $10^{-4}$ to $3{\cdot}10^{-3}$. GLIMPSE is stable over an order of magnitude of pretraining learning rates and diverges only at $10^{-1}$. Its selected rate of $3{\cdot}10^{-2}$ is high for a Transformer trained with AdamW; training at this rate is stable with 20 warmup epochs and gradient clipping at a norm of 1, and the cosine prediction loss is insensitive to the scale of the representations, whereas reconstruction alone prefers $10^{-3}$. Table~\ref{tab:weight-lr} shows that $\lambda_{\mathrm{pred}} = 0.3$ is the best weight at every learning rate from $3{\cdot}10^{-3}$ on. Table~\ref{tab:mask} shows that a masking ratio of 75\% is best for all three masked autoencoders, and also as the maximum of SMART's per-sample ratio, its original setting, although SMART is within 0.25 points from 25 to 90\%. Without masking, GLIMPSE-Rec has no target and its encoder remains untrained, and GLIMPSE, then trained by latent prediction alone, is not much better.

GLIMPSE-Pred, trained by latent prediction alone, prefers $3{\cdot}10^{-3}$, ten times lower than the rate of GLIMPSE, and the rate used in the ablation of Section~\ref{sec:results-ablation}. At the rate of GLIMPSE, $3{\cdot}10^{-2}$, it is 2.9 points lower and twice as variable across seeds (standard deviation of the mean over the eight tasks 1.4 against 0.7). With the same probe and targets, GLIMPSE and GLIMPSE-Rec reach 81.62 and 80.39 at their selected settings, so even tuned, latent prediction alone remains below reconstruction alone and below their combination.

In full finetuning, the encoder trains at $0.01\times$ the head learning rate, as in the finetuning configurations of DuETT and AID-MAE. STraTS and SMART instead train the whole model at one rate, as in their original recipes, because this was best for both: with the encoder at $0.01\times$, the best validation AUROC is 0.27 points lower for STraTS (82.45 against 82.72) and 0.03 points lower for SMART (82.36 against 82.39).

\begin{table}[htbp]
    \centering
    \small
    \caption{Learning rate selection: mean validation AUROC (\%) over the eight dataset--task combinations, three seeds. Pretraining rates are scored by a linear probe on the frozen representation (GLIMPSE with $\lambda_{\mathrm{pred}} = 0.3$). For full finetuning, the rate is that of the head; the encoder is trained at $0.01\times$ this rate, except for STraTS and SMART, whose whole model trains at this rate. Bold marks the selected rate of each method. STraTS, SMART and GLIMPSE-Pred (below the thin line in each part) were tuned later, with the reduced 16 PIC diagnosis targets, and numbers are not comparable with the ones above. GLIMPSE-Pred was not finetuned in this sweep.}
    \label{tab:lr}
    \begin{tabular}{lcccccccc}
        \toprule
        Method & $3{\cdot}10^{-5}$ & $10^{-4}$ & $3{\cdot}10^{-4}$ & $10^{-3}$ & $3{\cdot}10^{-3}$ & $10^{-2}$ & $3{\cdot}10^{-2}$ & $10^{-1}$ \\
        \midrule
        \multicolumn{9}{l}{\textit{Pretraining}} \\
        GLIMPSE-Rec & & & 77.96 & \textbf{78.80} & 78.52 & 77.95 & & \\
        GLIMPSE & & & & & 80.84 & 80.77 & \textbf{81.04} & 71.04 \\
        DuETT & & & 80.41 & 80.95 & \textbf{80.95} & 80.77 & & \\
        AID-MAE & & 77.07 & \textbf{77.67} & 77.48 & 77.41 & 76.61 & & \\
        \cmidrule(l){1-9}
        STraTS & & 79.02 & 79.53 & 79.93 & \textbf{80.18} & 78.91 & & \\
        SMART & & 81.07 & \textbf{81.23} & 81.14 & 81.21 & & & \\
        GLIMPSE-Pred & & & & 78.79 & \textbf{80.26} & 79.78 & 77.40 & \\
        \midrule
        \multicolumn{9}{l}{\textit{Full finetuning}} \\
        GLIMPSE-Rec & & & & 87.52 & 88.31 & \textbf{88.52} & 88.56 & \\
        GLIMPSE & & & & 88.08 & 88.65 & \textbf{88.80} & 88.83 & \\
        DuETT & 85.55 & 87.32 & \textbf{87.68} & 87.68 & & & & \\
        AID-MAE & & & 85.76 & 86.96 & 87.11 & 87.36 & \textbf{87.55} & 87.63 \\
        \cmidrule(l){1-9}
        STraTS & & 81.55 & 82.30 & 82.59 & \textbf{82.72} & 81.50 & & \\
        SMART & & & 82.20 & \textbf{82.39} & 82.33 & & & \\
        \bottomrule
    \end{tabular}
\end{table}

\begin{table}[htbp]
    \centering
    \small
    \caption{Weight $\lambda_{\mathrm{pred}}$ of the prediction loss and pretraining learning rate of GLIMPSE: mean validation AUROC (\%) of the linear probe, three seeds; the selected setting in bold.}
    \label{tab:weight-lr}
    \begin{tabular}{lccccc}
        \toprule
        $\lambda_{\mathrm{pred}}$ & $10^{-3}$ & $3{\cdot}10^{-3}$ & $10^{-2}$ & $3{\cdot}10^{-2}$ & $10^{-1}$ \\
        \midrule
        0.03 & & 80.23 & 79.97 & 79.35 & \\
        0.1 & 79.98 & 80.24 & 80.29 & 80.06 & \\
        0.3 & & 80.84 & 80.77 & \textbf{81.04} & 71.04 \\
        1 & & 80.33 & 80.38 & 79.97 & 70.63 \\
        \bottomrule
    \end{tabular}
\end{table}

\begin{table}[htbp]
    \centering
    \small
    \caption{Masking ratio: mean validation AUROC (\%) of the linear probe, three seeds, at each method's selected learning rate; the best ratio of each method in bold. SMART draws a ratio per sample uniformly from zero to the given maximum; it was tuned later (below the line) and is not comparable with the rows above.}
    \label{tab:mask}
    \begin{tabular}{lcccccc}
        \toprule
        Method & 0 & 0.1 & 0.25 & 0.5 & 0.75 & 0.9 \\
        \midrule
        GLIMPSE-Rec & 66.28 & & 77.34 & 77.98 & \textbf{78.75} & 78.05 \\
        GLIMPSE & 68.26 & & 80.47 & 80.66 & \textbf{81.14} & 79.95 \\
        AID-MAE & & 77.29 & & 78.12 & \textbf{78.78} & 77.91 \\
        \midrule
        SMART & & & 81.01 & 81.12 & \textbf{81.23} & 81.17 \\
        \bottomrule
    \end{tabular}
\end{table}

\subsection{Design of the Latent Prediction Objective}
\label{app:ablations}
The latent prediction objective involves several choices: what to predict, at which resolution and from which context lengths, with which loss and weight, and on which sample. We study them in the final protocol, pretraining on all training stays with $\lambda_{\mathrm{pred}} = 0.3$ and a learning rate of $3{\cdot}10^{-2}$, changing one choice at a time, with three seeds and both linear probing and full finetuning. Tables~\ref{tab:targets} and~\ref{tab:design} report the validation AUROC, on which every choice is made.

\paragraph{Prediction targets} We compare the following targets, where $s \in \{12, 24, 48\}$ hours is the context length:
\begin{itemize}
    \item \textbf{Forecast}: from the first half of a window of $2s$ hours, predict the representation of its second half.
    \item \textbf{Stay end}: from a context window of $s$ hours, predict the representation of the last $s$ hours of the stay.
    \item \textbf{Whole stay} (GLIMPSE): from a context window of $s$ hours anywhere in the stay, predict the representation of the whole stay.
    \item \textbf{Joint}: no separate reconstruction sample; a window of $s$ hours is masked and reconstructed, and the \texttt{[CLS]} output of this masked pass predicts the representation of the next $s$ hours.
    \item \textbf{CLS-JEPA}: the \texttt{[CLS]} output of the masked pass predicts the target encoder's \texttt{[CLS]} output of the masked values alone.
    \item \textbf{JEPA}: as in I-JEPA~\citep{assran2023ijepa}, a predictor maps the encoded visible values to the target encoder's representation of every masked value, using either a Transformer of the decoder's size of its own or the decoder itself.
\end{itemize}
Table~\ref{tab:targets} breaks the validation AUROC down by task type. The three targets that predict the representation of a window from another one (forecast, stay end and whole stay) are within 0.2 points of each other in the mean over the eight tasks, which is within the variability between seeds. They differ more between task types: forecast is ahead on length of stay, and the whole stay is ahead on diagnoses with linear probing, by 0.3 points, and tied for best with full finetuning. We use the whole stay because the diagnosis tasks are the most informative of the three: with 50 targets on MIMIC-IV and 16 on PIC, spanning the main organ systems, they probe a far broader range of conditions than a single mortality or length-of-stay label. The whole stay also requires no choice of horizon and gives every stay a single target. The gap opens where the target no longer extends beyond the context window. CLS-JEPA and JEPA predict the representation of masked values inside the window that the context encodes, and are 1.3 to 2.1 points below the best window target with linear probing and 0.7 to 1.0 points with full finetuning. The joint variants lie in between: their context is also a masked pass, but their target, the next $s$ hours, lies outside it, and they are 0.2 to 0.6 points below the best target with linear probing and 0.4 to 0.5 points with full finetuning. This seems to indicate that what matters is that the target reaches beyond the context window, not whether it covers the next hours, the last hours or the whole stay. With the last pretraining checkpoint instead of the best, the three window targets change by at most 0.3 points, so GLIMPSE uses the last checkpoint, which requires no validation criterion for pretraining.

\begin{table}[htbp]
    \centering
    \small
    \caption{Prediction targets: validation AUROC (\%), mean $\pm$ standard deviation over three seeds. Mortality and LOS average the three datasets, ICD averages MIMIC-IV and PIC, and Mean averages the eight dataset--task combinations. Forecast, stay end and whole stay use the pretraining checkpoint with the lowest validation reconstruction loss, the other targets the last checkpoint. Bold marks the best per column.}
    \label{tab:targets}
    \footnotesize
    \setlength{\tabcolsep}{4pt}
    \begin{tabular}{l*{8}{c}}
        \toprule
        & \multicolumn{4}{c}{Linear probing} & \multicolumn{4}{c}{Full finetuning} \\
        \cmidrule(lr){2-5} \cmidrule(lr){6-9}
        Target & \taskicon{mortality}Mortality & \taskicon{icd}ICD & \taskicon{los}LOS & Mean & \taskicon{mortality}Mortality & \taskicon{icd}ICD & \taskicon{los}LOS & Mean \\
        \midrule
        Forecast & 85.9\tiny{$\pm$0.4} & 82.1\tiny{$\pm$0.2} & \textbf{77.5}\tiny{$\pm$0.3} & \textbf{81.8}\tiny{$\pm$0.1} & 86.8\tiny{$\pm$0.3} & 83.4\tiny{$\pm$0.2} & \textbf{79.0}\tiny{$\pm$0.1} & 83.0\tiny{$\pm$0.2} \\
        Stay end & 85.8\tiny{$\pm$0.2} & 82.4\tiny{$\pm$0.1} & 77.1\tiny{$\pm$0.5} & 81.7\tiny{$\pm$0.2} & 86.8\tiny{$\pm$0.2} & \textbf{83.6}\tiny{$\pm$0.1} & 78.6\tiny{$\pm$0.3} & 82.9\tiny{$\pm$0.1} \\
        Whole stay & 85.7\tiny{$\pm$0.4} & \textbf{82.7}\tiny{$\pm$0.3} & 77.0\tiny{$\pm$0.3} & 81.7\tiny{$\pm$0.2} & \textbf{87.1}\tiny{$\pm$0.1} & \textbf{83.6}\tiny{$\pm$0.2} & 78.7\tiny{$\pm$0.1} & \textbf{83.1}\tiny{$\pm$0.0} \\
        Joint, 75\% masking & 85.2\tiny{$\pm$0.1} & 81.7\tiny{$\pm$0.2} & 76.9\tiny{$\pm$0.1} & 81.2\tiny{$\pm$0.1} & 86.7\tiny{$\pm$0.3} & 83.0\tiny{$\pm$0.3} & 78.4\tiny{$\pm$0.2} & 82.6\tiny{$\pm$0.2} \\
        Joint, 50\% masking & \textbf{86.0}\tiny{$\pm$0.2} & 81.9\tiny{$\pm$0.5} & 77.0\tiny{$\pm$0.7} & 81.6\tiny{$\pm$0.2} & 86.9\tiny{$\pm$0.3} & 83.1\tiny{$\pm$0.2} & 78.4\tiny{$\pm$0.2} & 82.7\tiny{$\pm$0.2} \\
        CLS-JEPA, 75\% masking & 84.5\tiny{$\pm$0.4} & 81.4\tiny{$\pm$0.2} & 75.8\tiny{$\pm$0.1} & 80.5\tiny{$\pm$0.2} & 86.4\tiny{$\pm$0.3} & 82.7\tiny{$\pm$0.3} & 78.3\tiny{$\pm$0.3} & 82.4\tiny{$\pm$0.1} \\
        CLS-JEPA, 50\% masking & 84.3\tiny{$\pm$0.3} & 80.8\tiny{$\pm$0.4} & 75.7\tiny{$\pm$0.2} & 80.2\tiny{$\pm$0.3} & 86.3\tiny{$\pm$0.3} & 82.6\tiny{$\pm$0.3} & 77.7\tiny{$\pm$0.1} & 82.2\tiny{$\pm$0.1} \\
        JEPA, own predictor, 75\% & 84.0\tiny{$\pm$0.4} & 81.1\tiny{$\pm$0.3} & 75.0\tiny{$\pm$0.2} & 79.9\tiny{$\pm$0.1} & 86.3\tiny{$\pm$0.3} & 82.6\tiny{$\pm$0.6} & 77.7\tiny{$\pm$0.3} & 82.2\tiny{$\pm$0.2} \\
        JEPA, own predictor, 50\% & 84.2\tiny{$\pm$0.6} & 80.9\tiny{$\pm$0.3} & 75.5\tiny{$\pm$0.1} & 80.1\tiny{$\pm$0.3} & 86.3\tiny{$\pm$0.1} & 82.7\tiny{$\pm$0.2} & 77.9\tiny{$\pm$0.2} & 82.3\tiny{$\pm$0.0} \\
        JEPA, decoder, 75\% & 83.6\tiny{$\pm$0.3} & 81.1\tiny{$\pm$0.1} & 75.0\tiny{$\pm$0.2} & 79.7\tiny{$\pm$0.1} & 86.0\tiny{$\pm$0.3} & 82.8\tiny{$\pm$0.3} & 77.6\tiny{$\pm$0.2} & 82.1\tiny{$\pm$0.1} \\
        JEPA, decoder, 50\% & 83.6\tiny{$\pm$0.4} & 81.0\tiny{$\pm$0.4} & 75.5\tiny{$\pm$0.2} & 79.9\tiny{$\pm$0.1} & 86.0\tiny{$\pm$0.5} & 82.7\tiny{$\pm$0.4} & 77.9\tiny{$\pm$0.1} & 82.2\tiny{$\pm$0.1} \\
        \bottomrule
    \end{tabular}
\end{table}

\paragraph{Resolution, context lengths, loss and sample} Table~\ref{tab:design} varies the remaining choices of the whole-stay objective. The number of bins per window, the context lengths and the form of the loss change the validation AUROC by at most 0.3 points. We use $K = 12$ bins, the cheapest resolution, and contexts of 12 to 48 hours. Only one change clearly hurts: reconstructing the context window itself instead of a separate sample of the stay, which loses 0.9 points in both settings. The reconstruction and prediction objectives thus benefit from seeing different parts of the stay.

\begin{table}[htbp]
    \centering
    \small
    \caption{Resolution, context lengths, loss and sample of the whole-stay objective: validation AUROC (\%), mean $\pm$ standard deviation over three seeds, last pretraining checkpoint. P12 averages its two tasks, MIMIC-IV and PIC their three, and Mean averages the eight dataset--task combinations. GLIMPSE uses $K = 12$, contexts of 12--48 hours, $\lambda_{\mathrm{pred}} = 0.3$, the cosine loss and a separate reconstruction sample. The context-length block is scored with the shared linear probe of the hyperparameter selection (Appendix~\ref{app:hyperparameters}) and not finetuned. Time is the mean pretraining time per run in minutes, over the three datasets and three seeds, each run on one node of four GH200 GPUs; the fastest in bold. Bold marks the best per column within each block.}
    \label{tab:design}
    \footnotesize
    \setlength{\tabcolsep}{4pt}
    \begin{tabular}{l*{8}{c}c}
        \toprule
        & \multicolumn{4}{c}{Linear probing} & \multicolumn{4}{c}{Full finetuning} & \\
        \cmidrule(lr){2-5} \cmidrule(lr){6-9}
        Variant & P12 & MIMIC-IV & PIC & Mean & P12 & MIMIC-IV & PIC & Mean & Time \\
        \midrule
        \multicolumn{10}{l}{\textit{Bins per window $K$ (contexts 12--48 h)}} \\
        $K=12$ (GLIMPSE) & \textbf{78.1}\tiny{$\pm$0.8} & 82.0\tiny{$\pm$0.4} & 83.7\tiny{$\pm$0.4} & \textbf{81.7}\tiny{$\pm$0.3} & 79.5\tiny{$\pm$0.5} & \textbf{83.4}\tiny{$\pm$0.3} & 84.8\tiny{$\pm$0.1} & 83.0\tiny{$\pm$0.2} & \textbf{56} \\
        $K=24$ & 77.5\tiny{$\pm$0.5} & 82.1\tiny{$\pm$0.3} & \textbf{83.8}\tiny{$\pm$0.4} & 81.6\tiny{$\pm$0.3} & 79.4\tiny{$\pm$0.5} & \textbf{83.4}\tiny{$\pm$0.1} & 84.9\tiny{$\pm$0.5} & 82.9\tiny{$\pm$0.2} & 72 \\
        $K=48$ & 77.9\tiny{$\pm$0.6} & \textbf{82.4}\tiny{$\pm$0.1} & 83.7\tiny{$\pm$0.8} & \textbf{81.7}\tiny{$\pm$0.4} & \textbf{79.6}\tiny{$\pm$0.7} & \textbf{83.4}\tiny{$\pm$0.3} & \textbf{85.1}\tiny{$\pm$0.6} & \textbf{83.1}\tiny{$\pm$0.4} & 111 \\
        \midrule
        \multicolumn{10}{l}{\textit{Context length}} \\
        12--48 h (GLIMPSE) & 78.0\tiny{$\pm$0.8} & 82.1\tiny{$\pm$0.4} & 83.6\tiny{$\pm$0.3} & 81.6\tiny{$\pm$0.3} & -- & -- & -- & -- \\
        6 h & 77.3\tiny{$\pm$0.4} & 81.5\tiny{$\pm$0.2} & 83.4\tiny{$\pm$0.2} & 81.2\tiny{$\pm$0.2} & -- & -- & -- & -- \\
        12 h & \textbf{78.5}\tiny{$\pm$0.8} & 81.6\tiny{$\pm$0.2} & 83.8\tiny{$\pm$0.3} & 81.6\tiny{$\pm$0.0} & -- & -- & -- & -- \\
        24 h & 78.0\tiny{$\pm$0.7} & 82.1\tiny{$\pm$0.2} & 83.5\tiny{$\pm$0.4} & 81.6\tiny{$\pm$0.1} & -- & -- & -- & -- \\
        48 h & 77.7\tiny{$\pm$0.4} & \textbf{82.2}\tiny{$\pm$0.3} & \textbf{83.9}\tiny{$\pm$0.4} & \textbf{81.7}\tiny{$\pm$0.3} & -- & -- & -- & -- \\
        \midrule
        \multicolumn{10}{l}{\textit{Prediction loss}} \\
        Cosine (GLIMPSE) & \textbf{78.1}\tiny{$\pm$0.8} & \textbf{82.0}\tiny{$\pm$0.4} & 83.7\tiny{$\pm$0.4} & \textbf{81.7}\tiny{$\pm$0.3} & \textbf{79.5}\tiny{$\pm$0.5} & \textbf{83.4}\tiny{$\pm$0.3} & 84.8\tiny{$\pm$0.1} & 83.0\tiny{$\pm$0.2} \\
        MSE & 77.7\tiny{$\pm$0.1} & 81.3\tiny{$\pm$0.3} & 84.3\tiny{$\pm$0.5} & 81.6\tiny{$\pm$0.1} & 79.0\tiny{$\pm$0.1} & 83.0\tiny{$\pm$0.2} & 85.5\tiny{$\pm$0.5} & 83.0\tiny{$\pm$0.1} \\
        L1 & 77.3\tiny{$\pm$0.4} & 81.6\tiny{$\pm$0.2} & \textbf{84.5}\tiny{$\pm$0.2} & 81.6\tiny{$\pm$0.2} & 79.3\tiny{$\pm$0.6} & 83.3\tiny{$\pm$0.1} & \textbf{85.7}\tiny{$\pm$0.2} & \textbf{83.2}\tiny{$\pm$0.1} \\
        Smooth L1 & 77.8\tiny{$\pm$0.8} & 81.9\tiny{$\pm$0.0} & 83.2\tiny{$\pm$0.2} & 81.4\tiny{$\pm$0.2} & \textbf{79.5}\tiny{$\pm$0.5} & \textbf{83.4}\tiny{$\pm$0.1} & 84.8\tiny{$\pm$0.1} & 83.0\tiny{$\pm$0.1} \\
        \midrule
        \multicolumn{10}{l}{\textit{Reconstruction sample}} \\
        Context window & 77.2\tiny{$\pm$0.6} & 81.7\tiny{$\pm$0.3} & 82.3\tiny{$\pm$0.4} & 80.8\tiny{$\pm$0.3} & 79.1\tiny{$\pm$0.8} & 82.7\tiny{$\pm$0.3} & 83.5\tiny{$\pm$0.0} & 82.1\tiny{$\pm$0.3} \\
        Up to 48 random bins (GLIMPSE) & \textbf{78.1}\tiny{$\pm$0.8} & \textbf{82.0}\tiny{$\pm$0.4} & \textbf{83.7}\tiny{$\pm$0.4} & \textbf{81.7}\tiny{$\pm$0.3} & \textbf{79.5}\tiny{$\pm$0.5} & \textbf{83.4}\tiny{$\pm$0.3} & \textbf{84.8}\tiny{$\pm$0.1} & \textbf{83.0}\tiny{$\pm$0.2} \\
        \bottomrule
    \end{tabular}
\end{table}

\paragraph{Comparison with I-JEPA} Latent prediction alone remains below reconstruction alone in GLIMPSE (Section~\ref{sec:results-ablation}). To check whether this is specific to our objective, we train a standard I-JEPA~\citep{assran2023ijepa} on the same tokens: no decoder and no \texttt{[CLS]} token, a Transformer predictor that predicts the target encoder's outputs at the masked values from the visible ones, and the mean of the target encoder's outputs as the representation. Table~\ref{tab:ijepa} sweeps its main settings, in the earlier protocol of the hyperparameter sweeps. The best setting remains 2.3 points below GLIMPSE on validation, which suggests that latent prediction on sparse ICU data benefits from being combined with reconstruction.

\begin{table}[htbp]
    \centering
    \small
    \caption{I-JEPA on the same tokens, one setting changed at a time: mean validation AUROC (\%) of the linear probe over the eight dataset--task combinations, three seeds, pretraining on the stays of at least 48 hours.}
    \label{tab:ijepa}
    \begin{tabular}{lc}
        \toprule
        Setting & Validation AUROC \\
        \midrule
        Default (LR $10^{-3}$, 75\% masking) & 73.0 \\
        LR $3{\cdot}10^{-4}$ & 71.4 \\
        LR $3{\cdot}10^{-3}$ & 77.6 \\
        LR $10^{-2}$ & 78.7 \\
        50\% masking & 73.0 \\
        90\% masking & 76.3 \\
        EMA 0.99 & 70.7 \\
        EMA 0.999 & 74.8 \\
        Multi-block masking & 72.3 \\
        Multi-block, LR $3{\cdot}10^{-4}$ & 69.9 \\
        Multi-block, LR $3{\cdot}10^{-3}$ & 74.8 \\
        \midrule
        GLIMPSE (whole stay) & 81.0 \\
        \bottomrule
    \end{tabular}
\end{table}

\subsection{Readout}
\label{app:readout}
The patient representation concatenates the \texttt{[CLS]} output and the mean of the value-token outputs (Section~\ref{sec:downstream}). The two pretraining objectives supervise different outputs of the encoder. The latent prediction objective acts on the \texttt{[CLS]} output alone and pushes it towards a summary of the whole stay, whereas the reconstruction loss, computed through the decoder from all encoder outputs, can be met by the value-token outputs, so value-level information need not be kept in the \texttt{[CLS]} output. A readout from the \texttt{[CLS]} token alone would lose this information, and a mean over the value tokens alone would lose the summary of the stay. Table~\ref{tab:readout} compares it with the \texttt{[CLS]} output alone and the mean alone, on the encoders of the main results. The concatenation is the best readout for linear probing with all three models, 0.8 to 0.9 points above the better of the two single readouts. With full finetuning, the three readouts are within 0.7 points of each other, with the \texttt{[CLS]} output slightly ahead.

\begin{table}[htbp]
    \centering
    \small
    \caption{Readout: mean validation AUROC (\%) over the eight dataset--task combinations, three seeds, on the encoders of the main results. Each cell gives linear probing / full finetuning; the best readout of each model and setting in bold.}
    \label{tab:readout}
    \begin{tabular}{lccc}
        \toprule
        Model & \texttt{[CLS]} & Mean & \texttt{[CLS]} + mean \\
        \midrule
        GLIMPSE-Rec & 78.3{\tiny $\pm$0.4} / \textbf{82.7}{\tiny $\pm$0.3} & 79.3{\tiny $\pm$0.2} / 82.5{\tiny $\pm$0.1} & \textbf{80.1}{\tiny $\pm$0.1} / 82.6{\tiny $\pm$0.1} \\
        GLIMPSE-Pred & 79.4{\tiny $\pm$0.9} / \textbf{82.5}{\tiny $\pm$0.3} & 78.0{\tiny $\pm$0.8} / 81.9{\tiny $\pm$0.2} & \textbf{80.3}{\tiny $\pm$0.6} / 82.4{\tiny $\pm$0.4} \\
        GLIMPSE & 80.7{\tiny $\pm$0.3} / \textbf{83.2}{\tiny $\pm$0.2} & 79.6{\tiny $\pm$0.6} / 82.6{\tiny $\pm$0.1} & \textbf{81.7}{\tiny $\pm$0.3} / 83.0{\tiny $\pm$0.2} \\
        \bottomrule
    \end{tabular}
\end{table}

\subsection{Compute}
\label{app:compute}
Every pretraining and finetuning run uses a single node with four NVIDIA GH200 GPUs (120\,GB each). One pretraining of 400 epochs takes about 20 minutes for GLIMPSE and GLIMPSE-Rec on PIC and P12 and 1.3 to 2.1 hours on MIMIC-IV; DuETT takes about 30 minutes on PIC and 3.4 to 4.3 hours on MIMIC-IV, and AID-MAE about 45 minutes on PIC and 2.7 to 2.9 hours on MIMIC-IV. One finetuning of 150 epochs takes 6 minutes at the median and at most one hour. The tabular foundation models run on a single GPU of the same type. Our models do not require a cluster: we were also able to pretrain and finetune them on a laptop (Lenovo Yoga with an Intel Core i9 CPU and an NVIDIA GeForce RTX 5060 Laptop GPU).

\clearpage
\section{Baselines}
\label{app:baselines}
\begin{figure}[htbp]
    \centering
    \begin{subfigure}[t]{0.45\textwidth}
        \includegraphics[width=\linewidth]{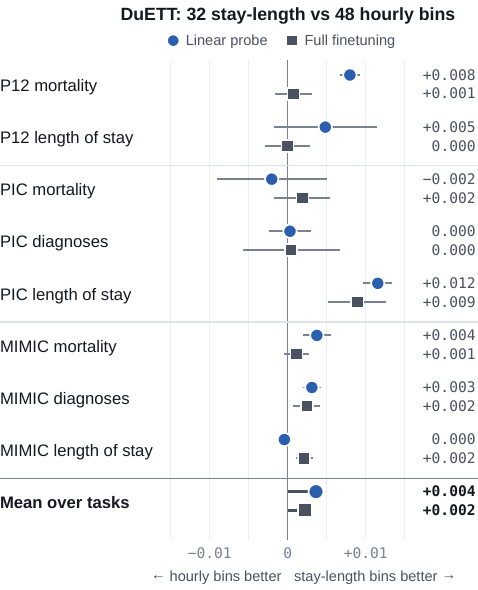}
        \caption{}
        \label{fig:duettbinning}
    \end{subfigure}\hfill
    \begin{subfigure}[t]{0.45\textwidth}
        \includegraphics[width=\linewidth]{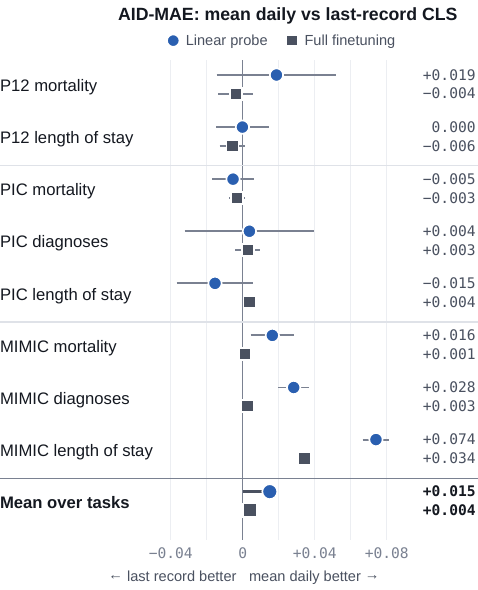}
        \caption{}
        \label{fig:aidmaecls}
    \end{subfigure}
    \caption{Validation AUROC of a baseline variant minus that of the alternative (on a 0--1 scale), per task and on average, three seeds; bars show the combined seed standard deviation. (a)~DuETT with its native stay-length bins minus hourly bins (MIMIC-IV: two seeds). (b)~AID-MAE with the mean \texttt{[CLS]} of the daily records minus the \texttt{[CLS]} of the last record.}
    \label{fig:variants}
\end{figure}

All baselines see the same stays, splits, labels and normalized values as GLIMPSE. The self-supervised encoders are pretrained on the full length of all training stays, like GLIMPSE, and evaluated with the protocol of Section~\ref{sec:exp-setting}: linear probing and full finetuning on the first 48 hours, fixed budgets of 400 pretraining and 150 finetuning epochs, learning rates selected on validation (Table~\ref{tab:lr}) and five seeds. Like GLIMPSE-Rec, every self-supervised baseline is finetuned from the pretraining checkpoint with the lowest validation loss of its own objective. For these objectives the validation loss decreases almost monotonically, so this checkpoint comes from late in training and is close to the last one. Only GLIMPSE uses the last checkpoint, because the validation loss of its latent prediction objective drops to its minimum early in training, before the representation is useful, so the checkpoint with the lowest validation loss would be one of the first (Appendix~\ref{app:hyperparameters}). Where a baseline has to be adapted to this protocol, we compare the adaptation with the alternatives on the validation set (Figure~\ref{fig:variants}).
 All four self-supervised baselines use the official implementations of their authors. DuETT's model code is taken unchanged from its public repository (commit \texttt{a99282c}). For AID-MAE, we use the code released with the paper (commit \texttt{0b5995e}), which has since been withdrawn from public access. We port its model into our training pipeline and verify with parity tests that the port reproduces the outputs of the released code, apart from the bug fix described below. The released AID-MAE code contains no finetuning implementation, so its downstream recipe follows the paper. For STraTS and SMART, we take the model code of their public repositories (commits \texttt{e936cda} and \texttt{f0c89f9}) and adapt only the input and the task head. SMART natively represents a stay as an hourly grid of values with an observation mask, so the input we have is essentially native; the only difference is that a bin keeps the latest value rather than the mean, as for every other model. STraTS natively embeds every raw measurement as a triplet of variable, exact timestamp and value. We keep the triplets but form them after the hourly binning, one per observed variable and hour with the bin's time since admission, so that STraTS sees the same values as the other models.
We keep the architectures of the original papers: DuETT embeds every variable and time bin in 24 dimensions and reads out one embedding per variable plus a summary token (5.6M encoder parameters), AID-MAE encodes daily records in 64 dimensions with an 8-layer encoder (416k parameters), STraTS embeds every observation triplet in 64 dimensions with a 2-layer encoder and combines them by attention into one 64-dimensional representation (83k parameters), and SMART embeds every variable in every hour in 32 dimensions with a 2-layer encoder and reads out one embedding per variable (38k parameters).

\subsection{DuETT\texorpdfstring{~\citep{labach2023duett}}{}}
\paragraph{Time binning} DuETT natively divides each stay into a fixed number of time bins (32 by default), so the duration of a bin depends on the length of the stay. We instead use bins of one hour, as for every other model, so that all encoders see the same temporal grid and differ only in their architecture and objective. Figure~\ref{fig:duettbinning} compares the two binnings. The native binning is slightly better on validation, by 0.4 AUROC points with linear probing and 0.2 points with full finetuning on average, less than half of the gap between DuETT and GLIMPSE in either setting. We keep the hourly bins so that the comparison isolates the architecture and the objective rather than the input format.

\subsection{AID-MAE\texorpdfstring{~\citep{xiang2026aidmae}}{}}
\paragraph{Representation of 48 hours} AID-MAE encodes a stay as a sequence of daily records, each summarized by its own \texttt{[CLS]} token, and the paper uses the first day for downstream tasks. Since every other model sees the first 48 hours, using only the first day would put AID-MAE at a disadvantage. We compare two ways to cover 48 hours:
\begin{itemize}
    \item \textbf{Daily average}: the mean of the \texttt{[CLS]} outputs of the first two daily records.
    \item \textbf{Last day}: the \texttt{[CLS]} output of the second day's record alone, which also refers to the first day through its historical reference values.
\end{itemize}
The daily average is better on validation, by 1.5 AUROC points with linear probing and 0.4 points with full finetuning (Figure~\ref{fig:aidmaecls}), and we use it throughout.

\paragraph{Attention of the \texttt{[CLS]} token to padding} AID-MAE removes missing and masked tokens before the encoder and right-pads each record to the longest record in the batch, so that, according to the paper, padded tokens receive zero attention. In the released code, however, the attention mask blocks the padding only for the data tokens: the CLS query still attends to every padded position. A record's CLS representation therefore depends on the other records in its batch, and on the batch size. We consider this an implementation bug and fix it with a one-line change to the attention mask, so that the CLS also attends only to observed tokens.

\paragraph{Masking ratio} The paper masks 25\% of the observed values. On our data, a ratio of 75\% is the best on validation (Table~\ref{tab:mask}), and we use it for AID-MAE as for GLIMPSE-Rec and GLIMPSE.

\subsection{Tabular Baselines}
\label{app:tabular}
\paragraph{Input}
The tabular baselines see the same stays, labels and values as our model: the first 48 hours of each stay, binned into 48 hourly bins from ICU admission (the latest value of each test in a bin, urine output summed), each test winsorized at its 5th and 95th percentile and standardized with the statistics of the training stays. A stay is a single row with one column per test and hourly bin, $48 \times T$ columns for $T$ tests, without aggregation and without a limit on the number of bins: 2{,}784 columns on PIC, 1{,}536 on P12 and 2{,}880 on MIMIC-IV. Columns that carry no information on the training stays (never measured, or constant) are removed, which removes a single column on P12.

\paragraph{Missing values}
Most cells are missing (94\% on PIC, 78\% on P12, 84\% on MIMIC-IV). XGBoost and TabPFN handle missing values natively, so they receive them as missing. Logistic regression cannot handle missing values. TabICLv2 and TabFM accept them but replace them by column means internally, without exposing which values were missing. For these three methods we impute missing values with the mean of the training stays and append one binary missingness indicator per column. Indicator columns that are identical on the training stays, as for tests measured together, are kept once: 1{,}611 indicators on PIC, 1{,}497 on P12 and 2{,}521 on MIMIC-IV.

\paragraph{Protocol}
Every binary label (in-hospital mortality, each ICD category) gets its own model, trained on the stays where the label is known; length of stay is a single multi-class model. No method uses class weighting. For logistic regression and XGBoost we select the hyperparameter configuration separately for every dataset, task and seed: the configuration with the highest validation AUROC (the macro average over the task's labels), as the encoders' finetuning selects on validation. The selected models are evaluated on the test set.

\paragraph{Logistic Regression}
We use scikit-learn's logistic regression with the liblinear solver (tolerance $10^{-3}$, at most 1{,}000 iterations), one-vs-rest for the multi-class length of stay. Following AID-MAE~\citep{xiang2026aidmae}, we search the penalty over $\{\ell_1, \ell_2\}$ and the inverse regularization strength $C$. We extend their grid $C \in \{0.1, 1, 10\}$ to $\{10^{-4}, 10^{-3}, 10^{-2}, 0.1, 1, 10\}$ (12 configurations), since the input has thousands of columns and calls for stronger regularization.

\paragraph{XGBoost}
We use XGBoost~\citep{chen2016xgboost} with the histogram method and otherwise default parameters, on AID-MAE's grid: learning rate $\in \{0.05, 0.1, 0.2\}$, maximum depth $\in \{4, 5, 6\}$ and $\{500, 1000\}$ trees (18 configurations). Missing values follow the default directions learned at each split.

\paragraph{Tabular foundation models}
The foundation models predict in context and are used as released, with their package's default settings and no hyperparameter search. All three fit the entire training split as context, without any cap on rows or columns, on a single NVIDIA GH200 GPU (120\,GB): up to 32{,}228 stays with 5{,}401 columns on MIMIC-IV.
\begin{itemize}
    \item \textbf{TabPFN}~\citep{hollmann2025tabpfn}: version 3.5 (package version 9.2.0), with its default ensemble. Its limits on the number of samples and features are the ranges it was pretrained for and can be exceeded, so we disable them.
    % TODO(tabular): cite the TabPFN 3.5 report if there is one
    \item \textbf{TabICLv2}~\citep{qu2026tabiclv2}: the released checkpoint (\texttt{tabicl-classifier-v2-20260212}) with the default ensemble of 8 members. To fit the MIMIC-IV context in memory, the members are processed one at a time and their column embeddings are offloaded to host memory, which does not change the predictions.
    \item \textbf{TabFM}~\citep{kong2026tabfmzeroshotfoundationmodel}: TabFM 1.1.0, the 410M-parameter model of the paper, with the default ensemble of 32 members, each of which sees a random subset of at most 500 columns, so that together they cover all columns.
\end{itemize}

\clearpage
\section{Extended Results}
\label{app:full-results}

\subsection{Label and Observation Efficiency}
\label{app:efficiency}
\label{app:label-efficiency}
\begin{figure*}[t]
    \centering
    \begin{subfigure}[t]{0.32\textwidth}
        \centering
        \includegraphics[width=\linewidth]{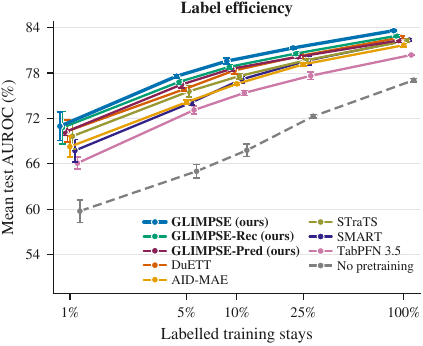}
        \caption{AUROC, full finetuning}
    \end{subfigure}\hfill
    \begin{subfigure}[t]{0.32\textwidth}
        \centering
        \includegraphics[width=\linewidth]{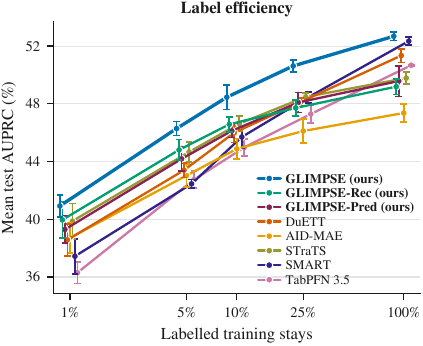}
        \caption{AUPRC, linear probing}
    \end{subfigure}\hfill
    \begin{subfigure}[t]{0.32\textwidth}
        \centering
        \includegraphics[width=\linewidth]{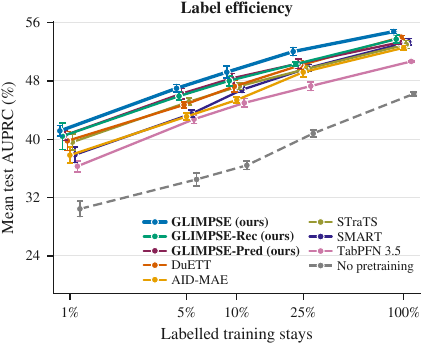}
        \caption{AUPRC, full finetuning}
    \end{subfigure}

    \vspace{0.5em}
    \begin{subfigure}[t]{0.32\textwidth}
        \centering
        \includegraphics[width=\linewidth]{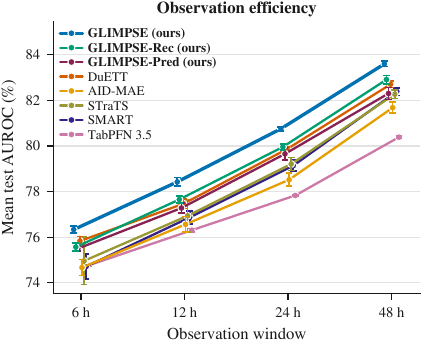}
        \caption{AUROC, full finetuning}
    \end{subfigure}\hfill
    \begin{subfigure}[t]{0.32\textwidth}
        \centering
        \includegraphics[width=\linewidth]{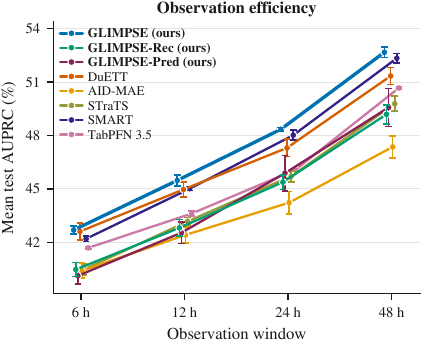}
        \caption{AUPRC, linear probing}
    \end{subfigure}\hfill
    \begin{subfigure}[t]{0.32\textwidth}
        \centering
        \includegraphics[width=\linewidth]{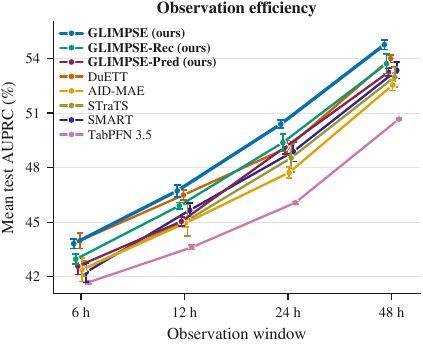}
        \caption{AUPRC, full finetuning}
    \end{subfigure}
    \caption{Label efficiency (a--c) and observation efficiency (d--f): AUROC in the setting not shown in Figure~\ref{fig:efficiency} (left) and AUPRC with linear probing and full finetuning (right). Mean over the eight tasks, bars show the standard deviation over 5 seeds; the 100\% and 48-hour points are Tables~\ref{tab:main} and~\ref{tab:main-auprc} (GLIMPSE-Pred: Table~\ref{tab:ablation}). Settings as in Figure~\ref{fig:efficiency}.}
    \label{fig:efficiency-appendix}
\end{figure*}

Figure~\ref{fig:efficiency-appendix} completes Figure~\ref{fig:efficiency} with the other setting and AUPRC, for the label efficiency (top row) and observation efficiency (bottom row) experiments. No pretraining: the GLIMPSE encoder with random weights, finetuned with the same recipe; it is only finetuned, so it appears in the full finetuning panels of label efficiency alone. In AUPRC, GLIMPSE is the best encoder at every label fraction in both settings, and at every observation window except with full finetuning at 6 hours, where DuETT is 0.2 points ahead (44.0 against 43.8). TabPFN~3.5 has no encoder, so its observation efficiency points are the same with linear probing and full finetuning.

Each seed draws its own subset of training stays, so the standard deviations include the choice of subset. At 1\%, some ICD targets have no positive stay in the subset: 4 to 8 of the 16 PIC targets, depending on the seed, and 1 or 2 of the 50 MIMIC-IV targets. The encoders are trained on them with negative labels only, and the tabular models predict the training prevalence for a target with fewer than two positive stays. Every target is still scored on the full validation and test splits, which contain positives, and kept in the macro average, so all methods are averaged over the same targets.

\subsection{Observation Windows on All Stays}
\label{app:windows}
\begin{figure*}[t]
    \centering
    \begin{subfigure}[t]{0.49\textwidth}
        \centering
        \includegraphics[width=\linewidth]{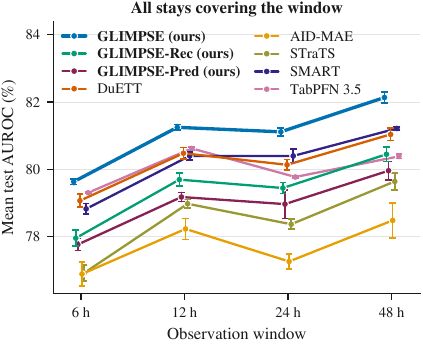}
        \caption{Linear probing}
    \end{subfigure}\hfill
    \begin{subfigure}[t]{0.49\textwidth}
        \centering
        \includegraphics[width=\linewidth]{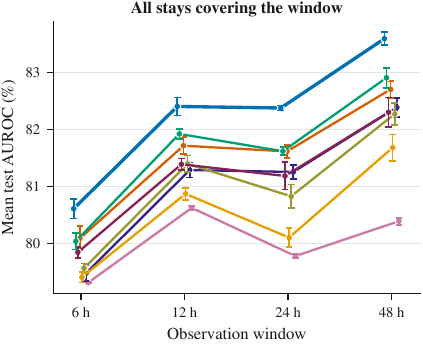}
        \caption{Full finetuning}
    \end{subfigure}
    \caption{Observation windows on all stays that cover the window: mean test AUROC over the eight tasks with a linear head on the frozen encoder (a) or full finetuning (b), on the stays of at least 6, 12, 24 or 48 hours, observed for that many hours. The cohort changes with the window, so methods should be compared within a window. Methods and bars as in Figure~\ref{fig:windows}.}
    \label{fig:windows-landmark}
\end{figure*}

Section~\ref{sec:results-windows} keeps the stays of at least 48 hours at every window, so that only the input changes. In clinical use, however, a prediction after $w$ hours is made for every patient still in the ICU, including those who leave before 48 hours. Figure~\ref{fig:windows-landmark} therefore repeats the experiment on all stays of at least $w$ hours, which adds shorter stays as the window shrinks. The encoders, the finetuning recipe and the normalization statistics are those of the main experiment. Two differences make the windows less comparable to one another than in Section~\ref{sec:results-windows}: the patients change with the window, and the length of stay task gains a third class, stays shorter than two days, at windows below 48 hours. The curves are therefore not monotonic, and methods should be compared within a window. The ranking matches the main experiment: GLIMPSE has the best mean AUROC at every window in both settings, ahead of DuETT by 0.6--1.1 points with linear probing and 0.5--0.9 points with full finetuning. TabPFN~3.5 is closer than in the main experiment, as the shorter stays are added: with linear probing, it is level with DuETT at 6 and 12 hours, 0.3 and 0.6 points behind GLIMPSE, and in AUPRC it is level with GLIMPSE at 6 hours (44.2 for both). With full finetuning, GLIMPSE leads it by 1.3--3.2 AUROC points at every window. At windows of at most 24 hours, AID-MAE sees a single daily record.

\subsection{Model Size}
\label{app:model-size}
\begin{figure*}[t]
    \centering
    \begin{subfigure}[t]{0.32\textwidth}
        \centering
        \includegraphics[width=\linewidth]{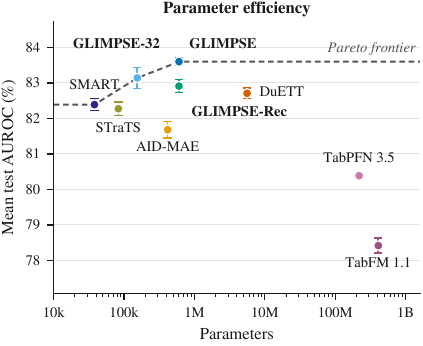}
        \caption{AUROC, full finetuning}
    \end{subfigure}\hfill
    \begin{subfigure}[t]{0.32\textwidth}
        \centering
        \includegraphics[width=\linewidth]{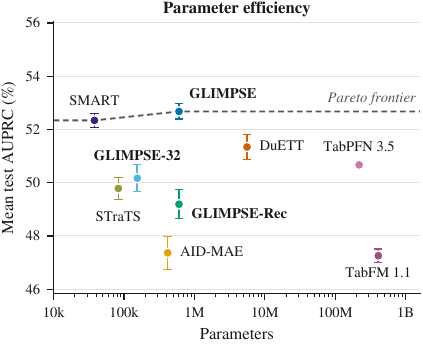}
        \caption{AUPRC, linear probing}
    \end{subfigure}\hfill
    \begin{subfigure}[t]{0.32\textwidth}
        \centering
        \includegraphics[width=\linewidth]{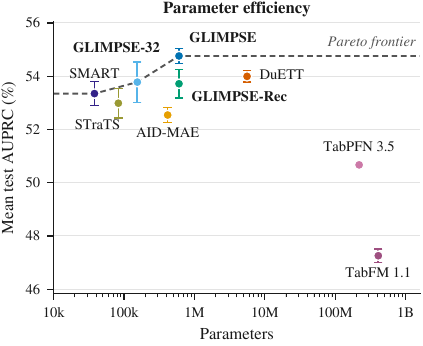}
        \caption{AUPRC, full finetuning}
    \end{subfigure}
    \caption{Parameter efficiency with full finetuning AUROC (a) and AUPRC with linear probing and full finetuning (b, c): mean test score and standard deviation over 5 seeds against the number of parameters (log scale). Models and parameter counts as in Figure~\ref{fig:cost}; the tabular foundation models are the same fits in every panel.}
    \label{fig:cost-full}
\end{figure*}

Table~\ref{tab:size} lists the number of parameters of every model of Figure~\ref{fig:cost} with its mean test AUROC and AUPRC. For the pretrained encoders, the count covers the encoder; the decoder, predictor and target encoder are used only in pretraining. GLIMPSE-32 is GLIMPSE with 32-dimensional embeddings, trained with the same settings. Figure~\ref{fig:cost-full} completes Figure~\ref{fig:cost} with full finetuning and AUPRC. In AUPRC, GLIMPSE is again the best model of any size in both settings; GLIMPSE-32 is 1.1 points behind DuETT with linear probing (50.2 against 51.3) and 0.2 points with full finetuning (53.8 against 54.0).

\begin{table}[htbp]
    \centering
    \small
    \caption{Number of parameters and mean test AUROC and AUPRC (\%) over the eight dataset--task combinations, five seeds. The tabular foundation models have no encoder to freeze and appear once. Bold marks the best per column.}
    \label{tab:size}
    \begin{tabular}{lccccc}
        \toprule
        & & \multicolumn{2}{c}{AUROC} & \multicolumn{2}{c}{AUPRC} \\
        \cmidrule(lr){3-4} \cmidrule(lr){5-6}
        Model & Parameters & Linear probing & Full finetuning & Linear probing & Full finetuning \\
        \midrule
        SMART & 38k & 81.2 & 82.4 & 52.3 & 53.3 \\
        STraTS & 83k & 79.6 & 82.3 & 49.8 & 53.0 \\
        GLIMPSE-32 (ours) & 154k & 81.0 & 83.1 & 50.2 & 53.8 \\
        AID-MAE & 416k & 78.5 & 81.7 & 47.4 & 52.5 \\
        GLIMPSE-Rec (ours) & 603k & 80.4 & 82.9 & 49.2 & 53.7 \\
        GLIMPSE (ours) & 603k & \textbf{82.1} & \textbf{83.6} & \textbf{52.7} & \textbf{54.8} \\
        DuETT & 5.6M & 81.0 & 82.7 & 51.3 & 54.0 \\
        \midrule
        TabICLv2 & 28M & \multicolumn{2}{c}{73.6} & \multicolumn{2}{c}{43.0} \\
        TabPFN 3.5 & 219M & \multicolumn{2}{c}{80.4} & \multicolumn{2}{c}{50.7} \\
        TabFM 1.1 & 410M & \multicolumn{2}{c}{78.4} & \multicolumn{2}{c}{47.3} \\
        \bottomrule
    \end{tabular}
\end{table}

\subsection{Comparison with Baselines in AUPRC}
\label{app:auprc}
Table~\ref{tab:main-auprc} repeats the comparison of Table~\ref{tab:main} in AUPRC, the AUPRC of the positive class of every label (Appendix~\ref{app:icd}), which weighs the positive stays more heavily than AUROC. The ranking is largely the same. With linear probing, GLIMPSE has the best mean AUPRC; SMART leads on both P12 tasks, and DuETT on MIMIC-IV diagnoses and on PIC mortality and diagnoses. With full finetuning, GLIMPSE again has the best mean and is the best model on PIC diagnoses and length of stay, while DuETT leads on PIC mortality. On P12 it is best on mortality and ties SMART on length of stay. On MIMIC-IV, GLIMPSE-Rec leads on mortality and length of stay and DuETT on diagnoses. The absolute values follow the prevalence of the positive class: they are low on the diagnosis tasks, where most targets are rare, and high for length of stay on P12 and PIC, where most stays last at least seven days.

\begin{table*}[t]
    \centering
    \footnotesize
    \setlength{\tabcolsep}{0.5pt}
    \caption{ Test AUPRC (\%), mean $\pm$ standard deviation over seeds; Mean averages the eight tasks. AUPRC of the positive class of every label; for length of stay, stays of at least seven days. Models and markers as in Table~\ref{tab:main}.}
    \label{tab:main-auprc}
        \begin{tabularx}{\textwidth}{l*{9}{>{\centering\arraybackslash}X}}
        \toprule
        & \multicolumn{2}{c}{P12} & \multicolumn{3}{c}{MIMIC-IV} & \multicolumn{3}{c}{PIC} & \\
        \cmidrule(lr){2-3} \cmidrule(lr){4-6} \cmidrule(lr){7-9}
        Model & \taskicon{mortality}Mortality & \taskicon{los}LOS & \taskicon{mortality}Mortality & \taskicon{icd}ICD & \taskicon{los}LOS & \taskicon{mortality}Mortality & \taskicon{icd}ICD & \taskicon{los}LOS & Mean \\
        \midrule
        \multicolumn{10}{l}{\textit{Tabular}} \\
        LogReg & 45.4\tiny{$\pm$0.5} & 85.9\tiny{$\pm$0.4} & 40.9\tiny{$\pm$0.3} & 30.5\tiny{$\pm$0.0} & 50.2\tiny{$\pm$0.2} & 29.1\tiny{$\pm$0.9} & 19.6\tiny{$\pm$0.3} & 77.2\tiny{$\pm$0.2} & 47.4\tiny{$\pm$0.2} \\
        XGBoost & 49.4\tiny{$\pm$1.6} & 86.2\tiny{$\pm$0.6} & 44.3\tiny{$\pm$0.5} & 32.5\tiny{$\pm$0.1} & 53.9\tiny{$\pm$0.4} & 25.2\tiny{$\pm$5.1} & 18.4\tiny{$\pm$0.3} & 78.2\tiny{$\pm$0.7} & 48.5\tiny{$\pm$0.6} \\
        TabPFN 3.5 & 52.1\tiny{$\pm$0.3} & 87.9\tiny{$\pm$0.2} & 43.3\tiny{$\pm$0.3} & 33.6\tiny{$\pm$0.0} & 55.4\tiny{$\pm$0.3} & 29.7\tiny{$\pm$1.3} & 23.1\tiny{$\pm$0.5} & 79.9\tiny{$\pm$0.3} & 50.6\tiny{$\pm$0.2} \\
        TabICLv2 & 45.1\tiny{$\pm$0.7} & 84.9\tiny{$\pm$0.5} & 31.9\tiny{$\pm$0.8} & 24.9\tiny{$\pm$0.5} & 43.7\tiny{$\pm$0.4} & 23.4\tiny{$\pm$3.3} & 14.5\tiny{$\pm$0.4} & 74.5\tiny{$\pm$0.7} & 42.9\tiny{$\pm$0.6} \\
        TabFM 1.1 & 47.1\tiny{$\pm$1.0} & 86.5\tiny{$\pm$0.4} & 39.6\tiny{$\pm$0.4} & 31.0\tiny{$\pm$0.3} & 50.0\tiny{$\pm$0.6} & 27.5\tiny{$\pm$1.6} & 19.6\tiny{$\pm$0.6} & 78.5\tiny{$\pm$1.0} & 47.5\tiny{$\pm$0.2} \\
        \midrule
        \multicolumn{10}{l}{\textit{Linear probing}} \\
        DuETT & 47.9\tiny{$\pm$1.3} & 85.6\tiny{$\pm$0.4} & 46.6\tiny{$\pm$0.7} & \underline{36.9}\tiny{$\pm$0.1} & 55.3\tiny{$\pm$0.2} & \underline{31.3}\tiny{$\pm$2.5} & \underline{27.6}\tiny{$\pm$1.4} & 81.6\tiny{$\pm$0.5} & 51.6\tiny{$\pm$0.5} \\
        AID-MAE & 38.8\tiny{$\pm$1.4} & 85.0\tiny{$\pm$0.8} & 40.3\tiny{$\pm$0.8} & 33.4\tiny{$\pm$0.3} & 50.4\tiny{$\pm$1.2} & 26.1\tiny{$\pm$3.7} & 24.2\tiny{$\pm$0.8} & 81.0\tiny{$\pm$0.8} & 47.4\tiny{$\pm$0.9} \\
        STraTS & 50.5\tiny{$\pm$1.2} & 87.1\tiny{$\pm$0.6} & 45.7\tiny{$\pm$1.1} & 33.0\tiny{$\pm$0.0} & 53.5\tiny{$\pm$0.4} & 28.1\tiny{$\pm$2.4} & 21.4\tiny{$\pm$0.4} & 79.6\tiny{$\pm$1.1} & 49.9\tiny{$\pm$0.3} \\
        SMART & \underline{54.0}\tiny{$\pm$1.1} & \underline{87.4}\tiny{$\pm$0.4} & 45.7\tiny{$\pm$0.4} & 36.3\tiny{$\pm$0.2} & 54.2\tiny{$\pm$0.6} & 30.0\tiny{$\pm$3.2} & 27.4\tiny{$\pm$1.9} & 80.8\tiny{$\pm$1.2} & 52.0\tiny{$\pm$0.7} \\
        GLIMPSE-Rec (ours) & 46.6\tiny{$\pm$2.5} & 86.3\tiny{$\pm$0.4} & 43.0\tiny{$\pm$0.5} & 33.9\tiny{$\pm$0.7} & 52.1\tiny{$\pm$1.3} & 28.6\tiny{$\pm$2.5} & 24.0\tiny{$\pm$1.5} & 79.1\tiny{$\pm$0.8} & 49.2\tiny{$\pm$0.6} \\
        GLIMPSE (ours) & 52.4\tiny{$\pm$1.5} & 87.1\tiny{$\pm$0.3} & \underline{46.7}\tiny{$\pm$1.2} & 36.3\tiny{$\pm$0.4} & \underline{56.3}\tiny{$\pm$0.3} & 29.2\tiny{$\pm$2.6} & 26.6\tiny{$\pm$0.9} & \underline{82.5}\tiny{$\pm$0.3} & \underline{52.1}\tiny{$\pm$0.3} \\
        \midrule
        \multicolumn{10}{l}{\textit{Full finetuning}} \\
        DuETT & 54.4\tiny{$\pm$1.0} & 86.9\tiny{$\pm$0.2} & 48.2\tiny{$\pm$1.1} & \textbf{38.7}\tiny{$\pm$0.2} & 58.6\tiny{$\pm$0.5} & \textbf{33.2}\tiny{$\pm$1.0} & 29.2\tiny{$\pm$1.1} & 83.1\tiny{$\pm$0.5} & 54.0\tiny{$\pm$0.3} \\
        AID-MAE & 52.5\tiny{$\pm$2.2} & 87.9\tiny{$\pm$0.8} & 48.6\tiny{$\pm$0.9} & 37.5\tiny{$\pm$0.2} & 58.9\tiny{$\pm$0.7} & 27.2\tiny{$\pm$1.2} & 25.6\tiny{$\pm$1.6} & 82.2\tiny{$\pm$0.3} & 52.6\tiny{$\pm$0.4} \\
        STraTS & 53.6\tiny{$\pm$2.3} & 88.5\tiny{$\pm$0.3} & 47.9\tiny{$\pm$1.2} & 37.1\tiny{$\pm$0.2} & 57.5\tiny{$\pm$0.8} & 29.5\tiny{$\pm$0.9} & 29.1\tiny{$\pm$1.6} & 82.3\tiny{$\pm$0.7} & 53.2\tiny{$\pm$0.3} \\
        SMART & 54.3\tiny{$\pm$0.8} & \textbf{88.9}\tiny{$\pm$0.4} & 48.4\tiny{$\pm$0.9} & 38.1\tiny{$\pm$0.1} & 57.6\tiny{$\pm$0.4} & 29.9\tiny{$\pm$0.6} & 26.8\tiny{$\pm$1.8} & 81.7\tiny{$\pm$0.9} & 53.2\tiny{$\pm$0.3} \\
        GLIMPSE-Rec (ours) & 52.8\tiny{$\pm$1.1} & 88.0\tiny{$\pm$0.3} & \textbf{49.9}\tiny{$\pm$0.9} & 38.5\tiny{$\pm$0.2} & \textbf{59.9}\tiny{$\pm$0.8} & 30.7\tiny{$\pm$1.3} & 26.6\tiny{$\pm$1.1} & 81.9\tiny{$\pm$0.6} & 53.5\tiny{$\pm$0.4} \\
        GLIMPSE (ours) & \textbf{55.8}\tiny{$\pm$1.0} & \textbf{88.9}\tiny{$\pm$0.2} & 49.3\tiny{$\pm$1.2} & 38.1\tiny{$\pm$0.4} & 59.1\tiny{$\pm$0.5} & 32.3\tiny{$\pm$2.2} & \textbf{29.3}\tiny{$\pm$1.5} & \textbf{84.1}\tiny{$\pm$0.6} & \textbf{54.6}\tiny{$\pm$0.3} \\
        \bottomrule
        \end{tabularx}
\end{table*}

\subsection{Reconstruction and Latent Prediction per Task}
\label{app:two-path}
Table~\ref{tab:ablation-tasks} breaks the ablation of Section~\ref{sec:results-ablation} down by task. With linear probing, combining both objectives gives the best result in every task. With full finetuning, GLIMPSE is best in five of the eight tasks; on MIMIC-IV, the largest cohort, where finetuning has the most labels to compensate for the pretraining, GLIMPSE-Rec is slightly ahead on mortality and diagnoses and GLIMPSE-Pred on length of stay, with the three variants within 0.5 points. GLIMPSE-Rec and GLIMPSE-Pred are close: with linear probing, GLIMPSE-Pred is ahead on length of stay on MIMIC-IV and PIC, level on P12 mortality, and behind on the other tasks.

\begin{table*}[t]
    \centering
    \footnotesize
    \setlength{\tabcolsep}{1pt}
    \caption{Ablation of the two pretraining objectives per task: test AUROC (\%), mean $\pm$ standard deviation over 5 seeds. Reconstruction only is GLIMPSE-Rec, prediction only GLIMPSE-Pred. Bold marks the best variant per column and setting.}
    \label{tab:ablation-tasks}
    \begin{tabularx}{\textwidth}{l*{9}{>{\centering\arraybackslash}X}}
        \toprule
        & \multicolumn{2}{c}{P12} & \multicolumn{3}{c}{MIMIC-IV} & \multicolumn{3}{c}{PIC} & \\
        \cmidrule(lr){2-3} \cmidrule(lr){4-6} \cmidrule(lr){7-9}
        Variant & \taskicon{mortality}Mortality & \taskicon{los}LOS & \taskicon{mortality}Mortality & \taskicon{icd}ICD & \taskicon{los}LOS & \taskicon{mortality}Mortality & \taskicon{icd}ICD & \taskicon{los}LOS & Mean \\
        \midrule
        \multicolumn{10}{l}{\textit{Linear probing}} \\
        GLIMPSE & \textbf{85.4}\tiny{$\pm$0.6} & \textbf{75.0}\tiny{$\pm$0.6} & \textbf{87.7}\tiny{$\pm$0.6} & \textbf{77.7}\tiny{$\pm$0.3} & \textbf{80.8}\tiny{$\pm$0.1} & \textbf{82.8}\tiny{$\pm$1.2} & \textbf{88.8}\tiny{$\pm$0.7} & \textbf{78.6}\tiny{$\pm$0.3} & \textbf{82.1}\tiny{$\pm$0.2} \\
        Reconstruction only & 83.4\tiny{$\pm$0.6} & 73.2\tiny{$\pm$0.5} & 86.7\tiny{$\pm$0.2} & 76.3\tiny{$\pm$0.5} & 78.2\tiny{$\pm$0.6} & 82.1\tiny{$\pm$1.0} & 86.7\tiny{$\pm$0.4} & 76.9\tiny{$\pm$0.5} & 80.4\tiny{$\pm$0.3} \\
        Prediction only & 83.4\tiny{$\pm$0.7} & 71.3\tiny{$\pm$1.2} & 85.4\tiny{$\pm$0.7} & 74.3\tiny{$\pm$0.8} & 79.1\tiny{$\pm$0.8} & 81.6\tiny{$\pm$1.0} & 86.1\tiny{$\pm$0.6} & 78.5\tiny{$\pm$0.7} & 80.0\tiny{$\pm$0.3} \\
        \midrule
        \multicolumn{10}{l}{\textit{Full finetuning}} \\
        GLIMPSE & \textbf{87.1}\tiny{$\pm$0.4} & \textbf{77.7}\tiny{$\pm$0.2} & 89.0\tiny{$\pm$0.3} & 78.8\tiny{$\pm$0.2} & 82.2\tiny{$\pm$0.3} & \textbf{84.4}\tiny{$\pm$0.9} & \textbf{89.6}\tiny{$\pm$1.0} & \textbf{80.5}\tiny{$\pm$0.7} & \textbf{83.7}\tiny{$\pm$0.2} \\
        Reconstruction only & 85.9\tiny{$\pm$0.3} & 75.8\tiny{$\pm$0.1} & \textbf{89.1}\tiny{$\pm$0.3} & \textbf{79.0}\tiny{$\pm$0.1} & 82.5\tiny{$\pm$0.3} & 82.7\tiny{$\pm$1.5} & 88.6\tiny{$\pm$0.4} & 78.8\tiny{$\pm$0.4} & 82.8\tiny{$\pm$0.2} \\
        Prediction only & 84.7\tiny{$\pm$0.3} & 74.4\tiny{$\pm$0.9} & 88.7\tiny{$\pm$0.2} & 78.7\tiny{$\pm$0.0} & \textbf{82.6}\tiny{$\pm$0.2} & 82.8\tiny{$\pm$0.6} & 87.6\tiny{$\pm$0.7} & 79.0\tiny{$\pm$0.8} & 82.3\tiny{$\pm$0.3} \\
        \bottomrule
    \end{tabularx}
\end{table*}

\end{document}